\documentclass{article}
\usepackage{ijcai23}

\usepackage{times}
\usepackage{soul}
\usepackage{url}
\usepackage[hidelinks]{hyperref}
\usepackage[utf8]{inputenc}
\usepackage[small]{caption}
\usepackage{graphicx}
\usepackage{amsmath}
\usepackage{amsthm}
\usepackage{booktabs}
\usepackage{algorithm}
\usepackage{algorithmic}
\usepackage[switch]{lineno}

\usepackage{color}
\usepackage{multirow}
\usepackage{amssymb}

\title{Multi-Modality Deep Network for JPEG Artifacts Reduction}

\author{
	Xuhao Jiang$^1$
	\and
	WeiminTan$^{1*}$\and
	Qing Lin$^1$\and
	Chenxi Ma$^1$ \and
	Bo Yan$^{1*}$  \And
	Liquan Shen$^2$  \\
	\affiliations
	$^1$School of Computer Science, Shanghai Key Laboratory of Intelligent Information Processing, Shanghai Collaborative Innovation Center of Intelligent Visual Computing, Fudan University, Shanghai, China\\
	$^2$School of Communication, Shanghai University, Shanghai, China
	\emails
	\{20110240011, wmtan, 18210240028, 17210240039, byan\}@fudan.edu.cn,
	 jsslq@163.com
	\thanks{Corresponding authors: Weimin Tan and Bo Yan. This work is supported by NSFC (Grant No.: U2001209, 61902076) and Natural Science Foundation of Shanghai (21ZR1406600).}	
}

\begin{document}

\maketitle

\begin{abstract}
In recent years, many convolutional neural network-based models are designed for JPEG artifacts reduction, and have achieved notable progress. However, few methods are suitable for extreme low-bitrate image compression artifacts reduction. The main challenge is that the highly compressed image loses too much information, resulting in reconstructing high-quality image difficultly. To address this issue, we propose a multimodal fusion learning method for text-guided JPEG artifacts reduction, in which the corresponding text description not only provides the potential prior information of the highly compressed image, but also serves as supplementary information to assist in image deblocking. We fuse image features and text semantic features from the global and local perspectives respectively, and design a contrastive loss built upon contrastive learning to produce visually pleasing results. Extensive experiments, including a user study, prove that our method can obtain better deblocking results compared to the state-of-the-art methods.
\end{abstract}

\section{Introduction}

Lossy image compression algorithms are widely used in image storage and transmission. However, due to the loss of information, complex compression noise is inevitably introduced into the compressed image, such as blocking artifacts \cite{dong2015compression}, resulting in degradation in both the visual quality of the compressed image and the performance of the subsequent computer vision tasks. Therefore, exploring methods for compressed image artifacts reduction is urgently needed, especially for the widely used JPEG format.

\begin{figure}[t]
	\begin{center}
		\includegraphics[width=1\linewidth]{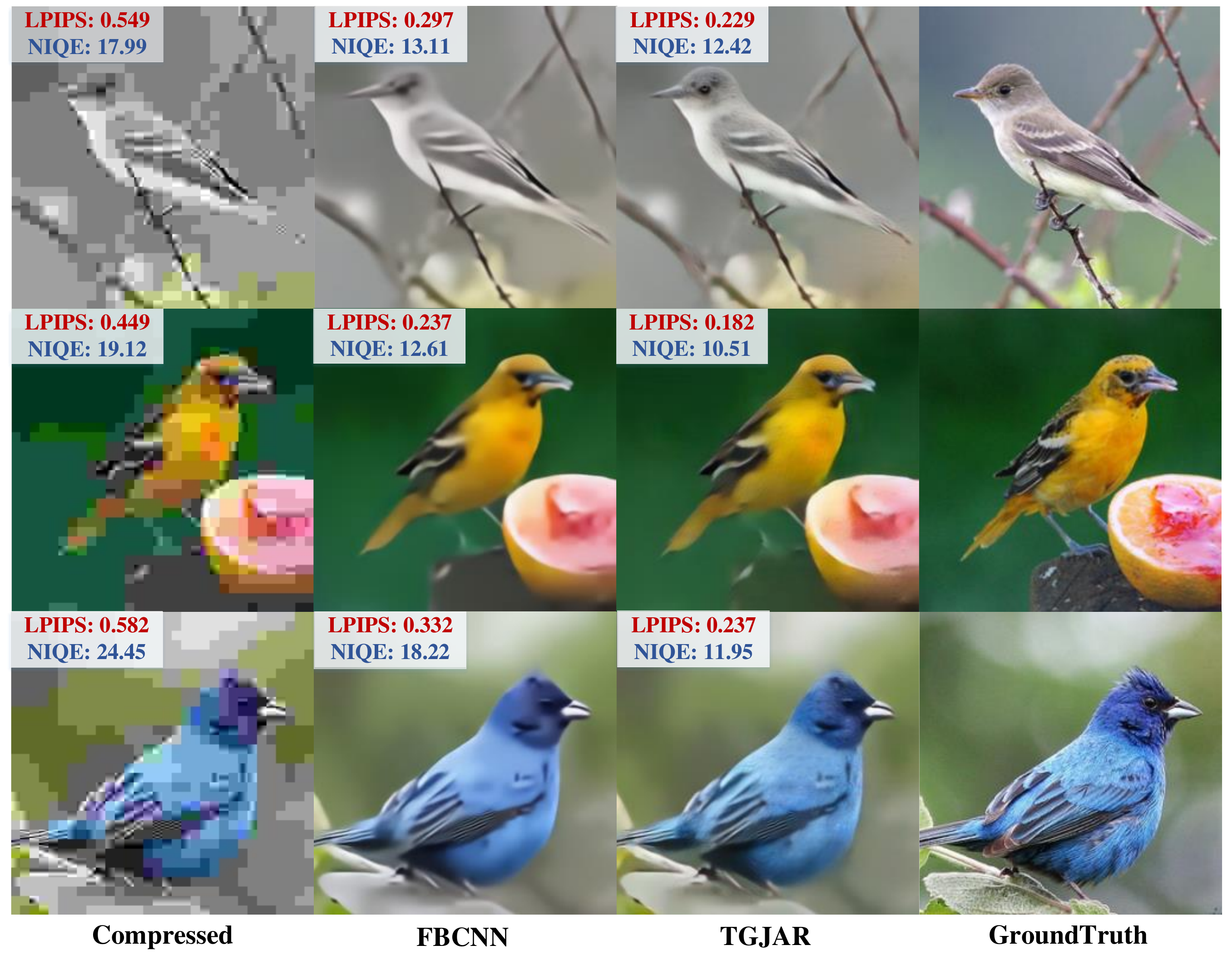}
	\end{center}
	\caption{Visual comparisons of the proposed TGJAR and the state-of-the-art (SOTA) method FBCNN at quality factor 1.}
	\label{fig:shouye}
\end{figure}

To cope with JPEG compression artifacts, many methods~\cite{zhang2017beyond,kim2019pseudo,jiang2021towards} have been proposed. However, in some occasions with limited bandwidth, the images are usually highly compressed for transmission, and the previous algorithms fail to effectively enhance the compressed image, as shown in Fig.~\ref{fig:shouye}. The main reason is that the highly compressed images lose too much information leading to the difficulty in image restoration. Faced with this extreme situation, JPEG artifacts reduction based on multimodal machine learning may have great advantages. The corresponding text provides the high-level image semantic information, which can be used as prior information to assist in image deblocking. Specifically, the text describes the main object of the image and some of its details, such as shape, color, location, etc. Under the guidance of the prior information, the multi-modality deep model can effectively remove the compression artifacts and reconstruct better deblocking results, as shown in Fig.~\ref{fig:shouye}.

In this paper, a text-guided JPEG artifacts reduction (TGJAR) generative adversarial network is proposed, in which text features including the word and sentence features are used to assist in image deblocking. This is an interesting attempt for multimodal machine learning. Considering that word and sentence features represent the local and global information of the text respectively, TGJAR proposes two kinds of image-text fusion modules to fuse image and text features from the global and local perspectives. For better local feature fusion, we adopt a multi-scale design to perform fusion on three different scales. Inspired by the contrast learning~\cite{Momentum2020He}, a contrastive loss is proposed to restrict the restored image from being pulled closer to the uncompressed image and away from the compressed image in the perceptual quality space. The main contributions are as follows:
\begin{itemize}
	\item We build a text-guided JPEG artifacts reduction (TGJAR) generative adversarial network, and design a multimodal fusion method to fuse compressed image features and the corresponding text features. To the best of our knowledge, we are the first to apply multimodal machine learning to image deblocking, and demonstrate the effectiveness of the text description guidance for JPEG artifacts reduction, especially for highly compressed images.

	\item Two kinds of fusion modules are employed to better fuse the features of text and images. The image-text global feature fusion can remove the global artifacts of the compressed image, and the local fusion modules employ an attention mechanism to obtain attention regions of word features for providing prior information.

	\item A well-designed contrastive loss is built upon contrastive learning to produce more realistic images.

	\item The experiments (including user study) show the outstanding perceptual performance of our TGJAR in comparison with the existing image deblocking methods.
	
\end{itemize}

\section{Related Work}

\textbf{JPEG Artifacts Reduction.} Recently, notable progress~\cite{fu2019jpeg,jin2020dual,liang2021swinir,zhang2020residual,chen2021enhanced,Fu2021TNNS,fu2021learning,zheng2019implicit,kim2020agarnet,zini2020deep,li2020learning,wang2022jpeg,jiang2022learning} has been made for JPEG artifacts reduction by utilizing deep convolutional neural networks. Dong \emph{et al.}~\cite{dong2015compression} first propose the famous ARCNN to solve this problem, which is a relatively shallow network. RNAN~\cite{zhang2019residual} designs local and non-local attention learning to further enhance the representation ability, and obtains good results in image restoration tasks, including image denoising, compression artifacts reduction, and image super-resolution. Wang \emph{et al.}~\cite{wang2021jpeg} proposes compression quality ranker-guided networks for JPEG artifacts reduction, which consist of a quality ranker network and a deblocking network. Li \emph{et al.} propose the QGCN~\cite{li2020learning} to handle a wide range of quality factors while it can consistently deliver superior image artifacts reduction performance. Jiang \emph{et al.}~\cite{jiang2021towards} propose a flexible blind CNN, namely FBCNN, which can predict the adjustable quality factor to control the trade-off between artifacts removal and details preservation. Witnessing the recent success of GAN in most image restoration tasks, some GAN-based JPEG artifacts reduction works~\cite{galteri2017deep,galteri2019deep} have been proposed, aiming to improve the subjective quality of compressed images. However, these methods show poor performance on recovering highly compressed image due to the serious loss of information. With the development of multimodal fusion learning technology, we can use the information of other modalities to assist in image deblocking.

\textbf{Multimodal Machine Learning.} The multimodal machine learning is a new and interesting topic, which simulates the cognitive process of humans by using information from multiple modalities. Some previous works~\cite{xu2018attngan,mittal2020emoticon,jiang2023multi} have demonstrated the powerful advantages of multimodal machine learning in the field of computer vision. For example, AttnGAN \cite{xu2018attngan} employs attention mechanism on the descriptive text to produce images with fine-grained details. Therefore, benefiting from the semantic information and the prior information provided by the text description, JPEG artifacts reduction based on multimodal machine learning may have a greater possibility to obtain better deblocking results.

\begin{figure*}[t]
	\begin{center}
		\includegraphics[width=0.9\linewidth]{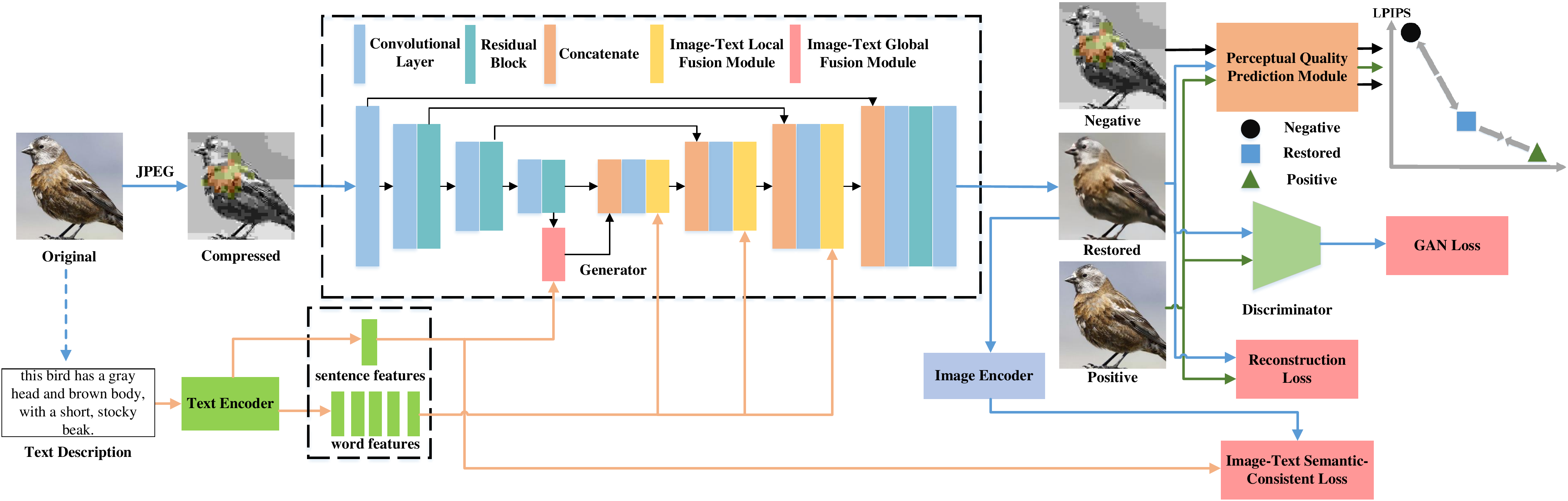}
	\end{center}
	\caption{The architecture of TGJAR. In our network, both the compressed image and the corresponding text description are taken as input. Specially, the text information is used as auxiliary information to reconstruct high perceptual quality images.}
	\label{fig:model}
\end{figure*}

\textbf{Contrastive Learning.} Recently, contrastive learning has demonstrated its effectiveness in self-supervised representation learning~\cite{Momentum2020He,simple2020chen}. The goal of the contrastive learning is to pull the target point toward the positive sample point and push the target point away from the negative sample point in a feature representation space. The contrastive learning enhances the contrast between positive and negative samples, which is beneficial for high-level vision tasks. For image pixel-level restoration, it is difficult to find a suitable feature space to construct positive and negative samples. In TGJAR, we propose to construct positive and negative samples in the image perceptual quality space, and design a novel contrastive loss for perceptual quality improvement.

\section{The Proposed TGJAR}
The process of compression algorithms can be expressed as
\begin{equation}
	I^c= F_c(I, QF)~,
\end{equation}
where $I^c$ is the compressed image, $I$ is the original uncompressed image, $F_c$ is the compression algorithm, and $QF$ represents the quality factor determining the degree of compression. The goal of JPEG artifacts reduction is to reconstruct a deblocking image $I^d$ from a compressed image $I^c$, aiming to keep $I^d$ and $I$ consistent in pixels.

When the image is highly compressed, the image information is severely lost, so it is difficult to recover a high-quality image through empirical modeling of the generator. In this scenario, we consider introducing other modal information (\emph{e.g.} text information) to assist highly compressed image deblocking. Our main goal is to remove JPEG artifacts in a highly compressed image with the assistance of a corresponding text description. The corresponding parameter optimization can be defined as below,
\begin{equation}
	\hat{\theta}_g= arg~\min_{\theta}\frac{1}{N}\sum_{n=1}^NL(G(I^c, T;{\theta}_g),I)~,
\end{equation}
where $G$ represents the generator, performing image artifacts reduction function, $L$ represents the loss function, $N$ represents the number of images in training dataset, and $T$ represents the text description. Based on this consideration, we propose a text-guided JPEG artifacts reduction generative adversarial network (TGJAR).

\subsection{Overview of our TGJAR}

The architecture design of the proposed TGJAR is shown in Fig.~\ref{fig:model}, which consists of five components: generator, discriminator, text encoder, image encoder and perceptual quality prediction module. Firstly, the text features, including word and sentence features, are extracted from the text description by using the text encoder. Then the sentence and word features are input into the generator to assist in image deblocking. Specially, we design the image-text local fusion module (LFM) and image-text global fusion module (GFM) in the generator to better fuse the compressed image features and text features. With the aid of text information, the generator can recover deblocking results with high perceptual quality. Specifically, the architecture of the generator is based on the U-Net \cite{U-Net} structure, and equipped with some residual blocks~\cite{he2016deep} in order to have a stronger deblocking ability. We also introduce the contrastive loss and the image-text semantic-consistent loss to further improve the perceptual quality of the deblocking results.

\subsection{Image-Text Fusion Module}

\begin{figure}[t]
	\begin{center}
		\includegraphics[width=1\linewidth]{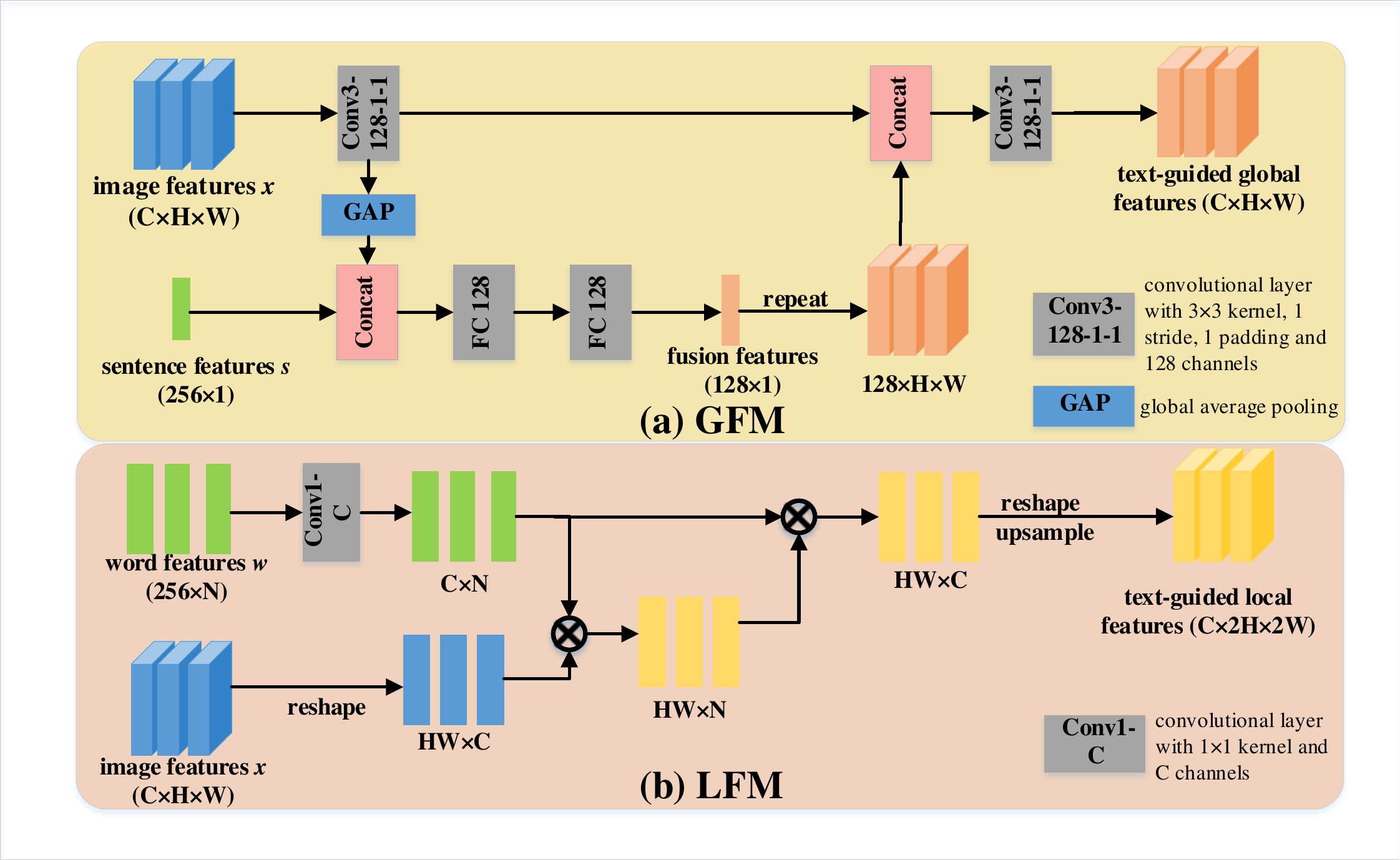}
	\end{center}
	\caption{The architecture of the proposed image-text global fusion module (a) and image-text local fusion module (b).}
	\label{fig:attention_model}
\end{figure}

In the TGJAR, the text encoder is a bi-direction Long Short-Term Memory (LSTM) \cite{Bidirectional1997Schuster}, which extracts the semantic features from the text description. Then we can obtain the word and sentence features respectively. Note that the word features represent the local features of a text description, while the sentence features denote the global features. In this way, the semantic information of text description can be mapped into a feature space consistent with image semantic information. The function of the text encoder can be defined as
\begin{equation}
	w, s=F_t(T)~,
\end{equation}
where $w$ and $s$ represent the word and sentence features respectively, $F_t$ represents the text encoder. Considering that the global and local features of the text can provide global and local prior information for the image respectively, two kinds of image-text fusion module including GFM and LFM, are designed to fuse the text and image features from both global and local perspectives. The architectures of the two modules are shown in Fig.~\ref{fig:attention_model}.

\textbf{Image-Text Global Fusion Module (GFM).} The architecture of GFM is shown in the Fig. \ref{fig:attention_model} (a). We directly use the features output by the third residual block in the generator to obtain the global image features. Given the $128\times32\times32$ feature map $x$, for global features, the feature map is further reduced to $128\times1\times1$ by employing a Conv3-128 and a global average pooling layer (GAP). The extracted image global features are concatenated with the sentence features $s$, and then two fully-connected (FC) layers are used to further extract the fused global features. The extracted $128\times1\times1$ fused global features are enlarged to the size of $128\times32\times32$ by repeating, and then concatenated with the input features. Finally, the concatenated features are processed by a Conv3-128 to get the text-guided global features.

Inspired by DPE~\cite{photo2018enhance}, GFM provides a way to fuse the global image features and the global text features, which utilizes the text-guided global features for image deblocking. With the aid of the text-guided global features, the image deblocking results achieve significant improvement on the overall image perceptual quality.

\textbf{Image-Text Local Fusion Module (LFM).} In TGJAR, we use three LFMs to adaptively fuse the local image features and the word features, which is based on a multi-scale strategy. The main goal of the multi-scale design is to improve the effect of word features in image deblocking task. The architecture of the LFM is shown in Fig.~\ref{fig:attention_model} (b). In the LFM, one $1\times1$ convolutional layer is used to adjust the size of the word features $w$, so as to calculate the correlation between the image features $x$ and the word features, and get the corresponding attention map. The input word features are weighted by the attention map, and then are reshaped and upsampled to obtain the text-guided local features.

LFM is designed to better fuse the image local features and the word features, which employs the attention mechanism to extract the corresponding attention regions of each word in compressed image. Then the attention regions in the compressed image can provide prior information for image deblocking. Thanks to the text-guided local features, the model removes the blocking artifacts and generates photorealistic details in the local region of the images, which makes the images look more real.

\subsection{Contrastive Learning}

Inspired by~\cite{Momentum2020He,simple2020chen}, we innovatively design a contrastive loss built upon contrastive learning to produce better results. The main difficulty lies in how to define the positive and negative samples and the corresponding feature constraint space. In TGJAR, we use compressed images as negative samples and uncompressed images as positive samples, and adopt the image perceptual quality space as the feature constraint space. The proposed contrastive loss constrains that the perceptual quality of the restored image is pulled to closer to that of the uncompressed image and pushed to far away from that of the compressed image. To the best of our knowledge, we are the first to propose a contrastive learning design constructed on the image quality assessment model. The details are as follows.

Here, we take the uncompressed image $I_i$ as the positive sample and the compressed image $I^c_i$ as the negative sample. The perceptual quality prediction module needs to be differentiable, so it can be any convolutional neural network-based model. We adopt the LPIPS \cite{zhang2018unreasonable} model in TGJAR, since LPIPS is highly consistent with human subjective evaluation. The image quality assessment (IQA) can be defined as $q_i=F_{lpips}(I^d_i,I_i),$ where $q_i$ represents the quality of the restored image $I^d_i$ with $I_i$ as a reference, and $F_{lpips}$ represents the function of LPIPS. Note that LPIPS is a reference-based IQA model. The predicted quality score is greater than or equal to zero, and lower score means that the perceptual quality of the predicted image is close to that of the reference image. Naturally, we can easily define the contrastive loss as

\begin{equation}
	L_{C}= ||\frac{F_{lpips}(I^d_i,I_i)-F_{lpips}(I_i,I_i)}{F_{lpips}(I^d_i,I_i)-F_{lpips}(I^c_i,I_i)}||_1,
\end{equation}

We can easily find that the $F_{lpips}(I_i,I_i)$ is equal to zero. Unfortunately, some related experiments prove that the above contrastive loss is not easy to converge, and even has a counterproductive effect. Thus, the $L_C$ is further simplified as
\begin{equation}
	L_{C}= \frac{F_{lpips}(I^d_i,I_i)}{F_{lpips}(I^d_i,I^c_i)+c},
\end{equation}
where $c$ is a constant. This simplified contrastive loss $L_{C}$ means that, taking $I_i$ as a reference, the perceptual quality value of $I^d_i$ should be small, and taking $I^c_i$ as a reference, the perceptual quality value of $I^d_i$ should be large. Its effectiveness has been verified in the experimental part.

\subsection{Loss Functions}

We use four different loss functions to optimize TGJAR, including contrastive loss $L_C$, reconstruction loss $L_R$, GAN loss $L_G$ and image-text semantic-consistent loss $L_{IT}$. The overall loss function is defined as
\begin{equation}
	L=\lambda_{1}L_{C}+\lambda_{2}L_{R}+\lambda_{3}L_{G}+\lambda_{4}L_{IT}
\end{equation}
where the $\lambda_{1}$, $\lambda_{2}$, $\lambda_{3}$, and $\lambda_{4}$ are the weights to balance different losses. $L_R$ and $L_G$ are commonly used in GAN-based models, and defined as, 
\begin{equation}
	L_{R}= ||I^d_i-I_i||_1 ,
\end{equation}
\begin{equation}
	\resizebox{0.9\linewidth}{!}{$
		L_{G}=E_{I_i\sim p_{data}(I)}[logD(I_i)] + E_{I^d_i\sim I^c}[log(1-D(I^d_i))]
		$}
\end{equation}
where the $D(.)$ represents the mapping funtion of the discriminator.

The image-text semantic-consistent loss is introduced to make the deblocking results semantic consistent with the text description. AttnGAN \cite{xu2018attngan} proposes a deep attentional multimodal similarity model to calculate the similarity of the text features and the features of the generated image. Following AttnGAN, an image encoder and a text encoder are employed to map the image features and text features into a common semantic space. The image encoder is based on the AttnGAN, which consists of two parts including a  Inception-v3~\cite{szegedy2016rethinking} model and a mapping layer. Then the image-text semantic-consistent loss is defined as
\begin{equation}
	L_{IT}=L_{word}+L_{sentence}
\end{equation}
where $L_{sentence}$ is the negative log posterior probability between $I^d_i$ and the corresponding sentence, and defined as $-(logP(s_i|I^d_i)+logP(I^d_i|s_i))$. Note that $w_i$ and $s_i$ are the word and sentence features of the corresponding text. Similarly, $L_{word}$ is the negative log posterior probability between the local region of $I^d_i$ and the corresponding words.

\begin{table*}[t]
	\scriptsize
	\begin{center}
		\begin{tabular}{c|c|c|c|c|c|c|c}
			\hline
			\multirow{1}*{Dataset} &\multirow{1}*{QF} &\multicolumn{1}{c|}{JPEG}&\multicolumn{1}{c|}{EDSR}&\multicolumn{1}{c|}{RNAN}&QGCN&FBCNN&\multicolumn{1}{c}{TGJAR}\\
			\cline{1-8}
			\hline
			\multirow{3}*{CUB}&1&0.505/202.5/11.45/19.43&0.349/92.8/8.17/10.36&0.356/93.4/7.89/9.74&0.341/90.8/8.54/11.00&0.332/74.2/8.56/11.06&\textbf{0.249/17.9/6.95/8.62}\\
			\cline{2-8}			
			~&5&0.375/116.3/~8.87~/14.39&0.248/59.6/7.19/~9.34~&0.253/54.3/6.99/8.81&0.239/53.9/7.56/~9.91~&0.237/47.5/7.58/~9.84~&\textbf{0.154/10.7/6.12/8.08}\\
			\cline{2-8}		
			~&10&0.228/~42.9~/~6.72~/10.32&0.156/34.3/6.45/~8.71~&0.161/36.9/6.08/8.14&0.154/33.0/6.72/~9.09~&0.155/31.2/6.65/~8.90~&\textbf{0.089/~6.9~/5.60/7.69}\\
			\cline{1-8}	
			\hline
			\multirow{3}{*}{\begin{tabular}[c]{@{}c@{}}Oxford\\ -102\end{tabular}}&1&0.493/222.3/10.68/18.64&0.309/74.1/7.91/9.70&0.315/75.7/7.62/9.14&0.294/72.1/7.97/10.04&0.290/66.6/7.94/9.98&\textbf{0.224/42.0/6.34/7.83}\\
			\cline{2-8}				
			~&5&0.342/126.6/~8.18~/13.29&0.201/60.2/6.90/8.49&0.206/59.3/6.76/8.14&0.190/58.5/6.96/~8.77~&0.190/53.8/6.97/8.70&\textbf{0.142/32.2/5.70/7.36}\\
			\cline{2-8}		
			~&10&0.185/~60.3~/~6.21~/~9.37~&0.125/48.1/6.27/7.85&0.128/47.6/6.15/7.68&0.116/46.5/6.28/~8.08~&0.120/42.4/6.35/8.08&\textbf{0.080/25.4/5.53/7.11}\\
			\cline{1-8}
			\hline
		\end{tabular}
		\caption{Average LPIPS$|$FID$|$PI$|$NIQE values of various methods based on the color images from CUB and Oxford-102 datasets for QF = 1, 5 and 10. Lower is better. The best results are boldfaced.
		}\label{tab:lpips}
	\end{center}
	
\end{table*}

\begin{table}[t]
	\scriptsize
	\begin{center}
		\begin{tabular}{c|c|c|c|c|c|c|c}
			\hline
			\multirow{1}*{Dataset} &\multirow{1}*{QF} &\multicolumn{1}{c|}{JPEG}&\multicolumn{1}{c|}{EDSR}&\multicolumn{1}{c|}{RNAN}&QGCN&FBCNN&\multicolumn{1}{c}{TGJAR}\\
			\cline{1-8}
			\hline
			\multirow{3}*{CUB}&1&21.8&24.4&24.3&24.6&24.7&24.4\\
			\cline{2-8}			
			~&5&24.3&26.9&26.7&27.2&27.2&26.8\\
			\cline{2-8}		
			~&10&27.4&29.8&29.6&30.0&29.9&29.7\\
			\cline{1-8}	
			\hline
			\multirow{3}{*}{\begin{tabular}[c]{@{}c@{}}Oxford\\ -102\end{tabular}}&1&21.3&24.0&23.9&24.2&24.2&24.0\\
			\cline{2-8}				
			~&5&23.7&26.6&26.4&26.9&26.9&26.5\\
			\cline{2-8}		
			~&10&26.8&29.6&29.4&30.0&29.8&29.5\\
			\cline{1-8}
			\hline
		\end{tabular}
		\caption{Average PSNR values of various methods based on the color images from CUB and Oxford-102 datasets for QF = 1, 5 and 10. Higher is better. 
		}\label{tab:psnr}
	\end{center}
	
\end{table}

\subsection{Training Implementation}
In the proposed TGJAR, training is divided into two stages. In the first stage, following AttnGAN~\cite{xu2018attngan}, the image encoder and text encoder are pretrained, aiming to map the image features and text features into a common semantic space. In the second stage, we fix the weights of the image encoder, text encoder and the perceptual quality prediction module, and train the discriminator and generator. Note that the second stage is our main contribution. In the second training stage, Pytorch is used as the training toolbox, and the Adam optimization algorithm~\cite{Adam} with a mini-batch of 4 is adopted for training. All the experiments are conducted on a NVIDIA GeForce RTX 1080 Ti. The learning rate is changed from $1\times10^{-4}$ to $1\times10^{-8}$ at the interval of twenty epochs. The hyper-parameter $c$ of the $L_C$ is set as 0.1, and the hyper-parameters $\lambda_{1}$, $\lambda_{2}$, $\lambda_{3}$, and $\lambda_{4}$ of the global loss function are empirically set as 0.01, 1, 0.001 and 0.0005, respectively.

\section{Experimental Results}

\subsection{Datasets and Evaluation Methodology}
We evaluate our TGJAR on the CUB~\cite{CUB} and Oxford-102~\cite{flower} datasets, in which all images are annotated with corresponding text descriptions. CUB dataset contains 200 species of bird with a total of 11,788 images, of which 8,855 images are used for training and 2,933 images are used for testing. Oxford-102 Dataset consists of 102 flower categories, with a total of 8,189 images including 7,034 images for training and 1,155 images for testing. We preprocess the two datasets according to the methods in AttnGAN \cite{xu2018attngan}, then crop and resize the images into patches of size $256\times256$.

Considering TGJAR aims at improving the perceptual quality of the highly compressed image, we adopt small QFs (i.e., 1, 5 and 10) of JPEG compression algorithm to process the training and testing datasets. Following~\cite{PI,mentzer2020high}, we evaluate the proposed TGJAR and the compared methods in four perceptual quality metrics, including LPIPS~\cite{zhang2018unreasonable}, FID~\cite{FID}, PI~\cite{PI} and NIQE~\cite{NIQE}, which are highly consistent with human perception of images. Besides, we use Peak Signal-to-Noise Ratio (PSNR) to measure the fidelity of the reconstructions.

\subsection{Comparison with The SOTA Methods}
In this part, TGJAR and the SOTA algorithms including EDSR~\cite{EDSR}, RNAN~\cite{zhang2019residual}, QGCN~\cite{li2020learning} and FBCNN~\cite{jiang2021towards} are compared quantitatively and qualitatively. To conduct a fair comparison, EDSR and RNAN are retrained on these two datasets including CUB and Oxford-102. In particular, we remove the upsampling module and set the number of the residual block to 16 in EDSR, and use the
RGB compressed images to train RNAN. For QGCN and FBCNN, they are finetuned on the training datasets since the pretrained models are available. Since the training codes for the existing GAN-based models are not available, TGJAR is not compared with them. Certainly, we compare the proposed TGJAR with our baseline model (i.e., a GAN-based model) in the ablation experiments.

\begin{figure}[t]
	\begin{center}
		\includegraphics[width=1\linewidth]{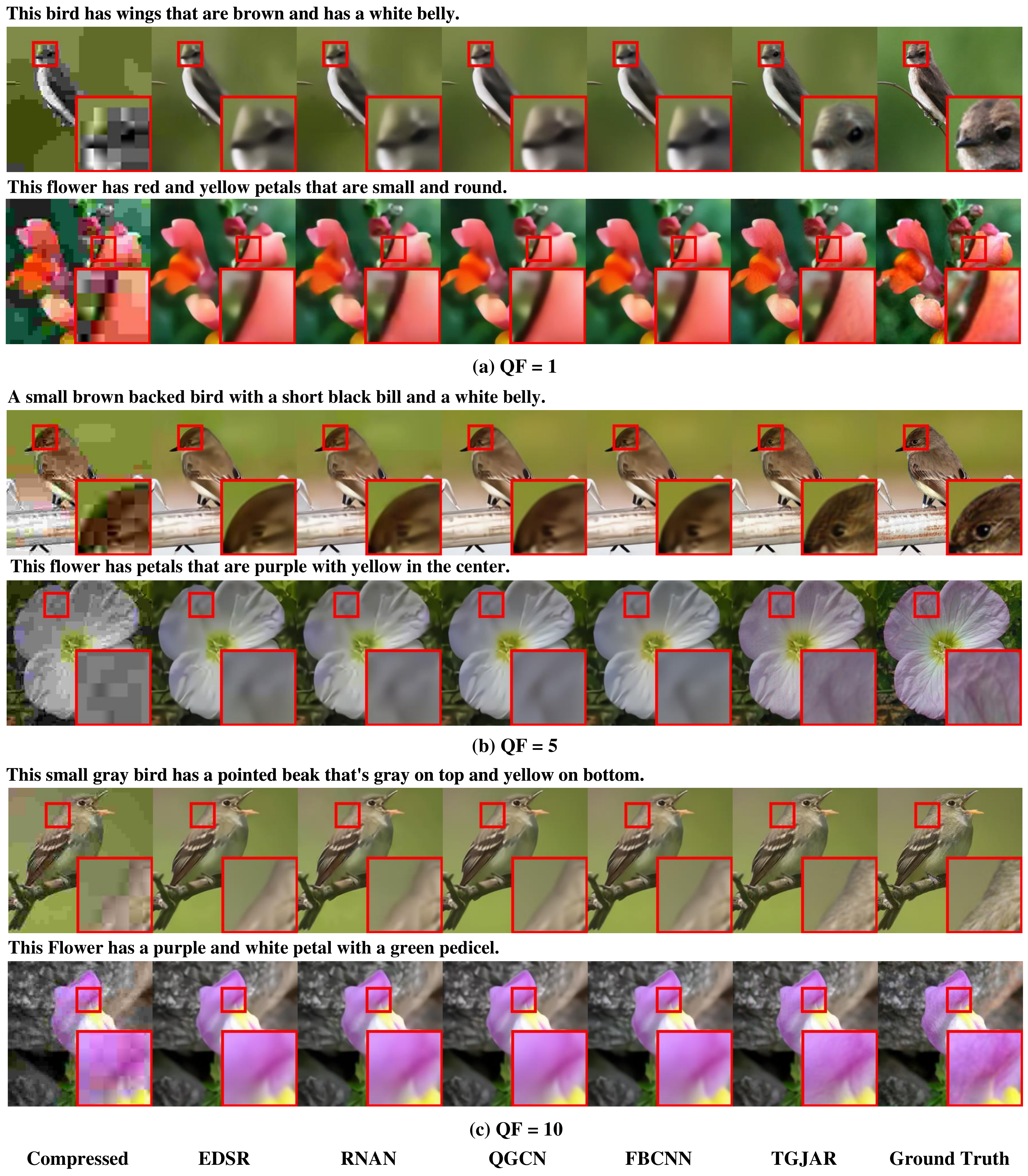}
	\end{center}
	\caption{Visual comparisons with SOTA methods on CUB and Oxford-102 datasets. Above each line of the images is the corresponding text description. Better zoom in.}
	\label{fig:bird}
\end{figure}

\textbf{Quantitative Comparisons.} Tables \ref{tab:lpips} and \ref{tab:psnr} show the quantitative results on two datasets with JPEG QF 1, 5 and 10, respectively. In Table \ref{tab:lpips}, the proposed TGJAR achieves the best performance on the four perceptual quality indexes (i.e. LPIPS, FID, PI and NIQE) at all JPEG QFs. Specially, we notice that our deblocking results of FID on two datasets at QF 1 are better than that of other methods on two datasets at QF 10. As shown in Table \ref{tab:psnr}, it can be found that the deblocking results of these methods all achieve significant PSNR gain compared with the compressed images. Specifically, the proposed TGJAR achieves a competitive performance on PSNR compared to other four methods. EDSR, RNAN, QGCN and FBCNN are all MSE-based JPEG artifacts reduction methods, which usually produce overly smooth deblocking results. This results in good performance on PSNR, but poor performances on the other four perceptual quality indexes which are more consistent with human perception. Compared with them, our proposed TGJAR obtains competitive PSNR, and achieves significant improvement on LPIPS, FID, PI and NIQE. The main reason is that TGJAR uses text description as auxiliary information to improve the perceptual qualities of deblocking results, where text not only provides prior information of the compressed image, but also supplements semantic information losses. From Tables \ref{tab:lpips} and \ref{tab:psnr}, we can find that TGJAR takes into account the
faithfulness of the reconstruction to the uncompressed image, and greatly improves the perceptual quality of the deblocking results compared with existing algorithms.

\begin{figure}[t]
	\begin{center}
		\includegraphics[width=0.95\linewidth]{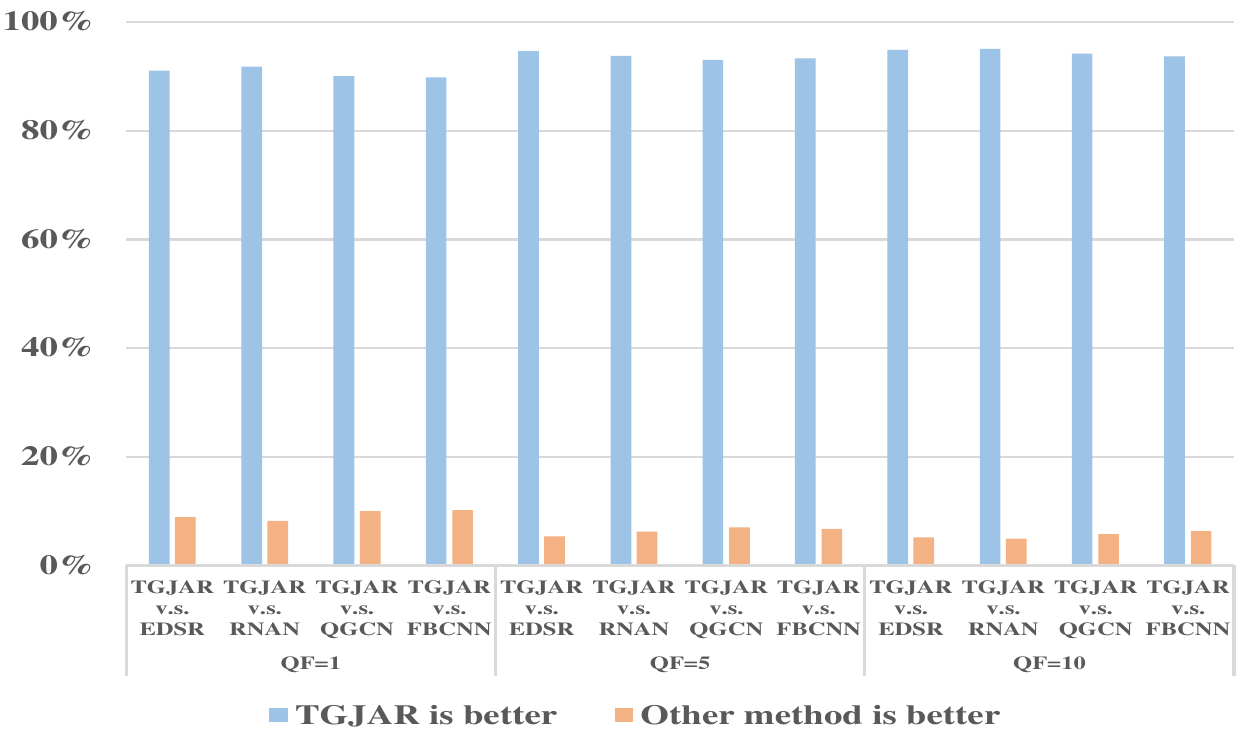}
	\end{center}
	\caption{User study results. The reported value indicates the performance rate of the proposed TGJAR against the other methods at QF=1, 5 and 10, respectively.}
	\label{fig:user}
\end{figure}

\textbf{Qualitative Comparisons.} The proposed TGJAR not only outperforms the comparative methods in terms of overall quantitative evaluation, but also produces deblocking results containing high subjective quality. As shown in Fig.~\ref{fig:bird}, four deblcocking methods all remove the JPEG artifacts to a large extent. However, the deblocking results of EDSR, RNAN, QGCN and FBCNN seem blurry, especially at QF=1. The main reason is that under a high compression rate, the image information is seriously lost resulting in reconstructing high-quality deblocking results difficultly. Unlike these four methods, TGJAR makes full use of text information, and generates appealing visual results containing clear edges and rich textures even in the case of QF=1.

\subsection{User Study}

We further conduct a user study with 13 subjects. Given a pair of images produced by two methods, the users are asked to judge which one owns higher perceptual quality and is more coherent with the given texts. We randomly select 50 images in each dataset at each QF value, and this user study requires 15,600 comparisons in total. The results are shown in Fig.~\ref{fig:user}. 
We can find that our results win more than 90\%. This subjective comparisons are consistent with the quantitative comparisons in Table~\ref{tab:lpips}, which demonstrate that TGJAR outperforms other methods.

\subsection{Ablation Study}

\begin{table}[t]
	\begin{center}
		\resizebox{1\linewidth}{!}{
		\begin{tabular}{ccccc|cc}
			\hline
			\multicolumn{5}{c|}{Attributes} &\multicolumn{2}{c}{Quality Index}\\
			\cline{1-7}
			\multirow{2}*{$L_R$}&\multirow{2}*{$L_G$}&\multirow{1}*{$GFM$}&\multirow{1}*{$LFM$}&\multirow{2}*{$L_C$}&\multirow{2}*{PSNR$\uparrow$}&\multirow{2}*{LPIPS/FID/PI/NIQE$\downarrow$}\\
			~&~&+$L_{IT}$&+$L_{IT}$&~&~&~\\
			\hline
			\checkmark&\checkmark&&&~&\textbf{24.4}&0.342/79.7/8.57/11.04\\
			\checkmark&\checkmark&\checkmark&&~&\textbf{24.4}&0.314/28.4/8.43/10.93\\
			\checkmark&\checkmark&&\checkmark&~&24.2&0.307/36.7/8.44/11.05\\
			\checkmark&\checkmark&\checkmark&\checkmark&&24.3&0.308/33.3/8.36/10.86\\
			\checkmark&\checkmark&\checkmark&\checkmark&\checkmark&\textbf{24.4}&\textbf{0.249/17.9/6.95/~8.62~~}\\
			\hline
		\end{tabular}
	}
	\end{center}
	\caption{Performance comparisons between variations of our TGJAR on CUB at QF=1. The best results are boldfaced.}\label{tab:ablation}
\end{table}

We conduct the ablation study to verify the effectiveness of the proposed contrastive loss and two image-text fusion modules including GFM and LFM. Note that TGJAR only optimized by $L_R$ and $L_G$ is regarded as the baseline model. Considering that the $L_{IT}$ can constrain deblocking results more consistent with the text description, we employ the $L_{IT}$ when using the image-text fusion modules. This allows the TGJAR to make better use of text information.

The quantitative comparisons at QF 1 are shown in Table \ref{tab:ablation}. It can be found that all variations of TGJAR obtain similar performance on PSNR index. This demonstrates that TGJAR still faithfully reconstructs the deblocking results to the uncompressed images after introducing text information. On the other four indicators, the final model achieves the best performance. Among them, the model with GFM or LFM shows similar performance. Compared with these two models, the model with both GFM and LFM shows intermediate performance on LPIPS and FID, and better performance on PI and NIQE, since LFM and GFM can complement each other. Finally, the contrastive loss improves the performance of the model on four indicators by simultaneously using positive and negative samples.

\begin{figure}[t]
	\begin{center}
		\includegraphics[width=1\linewidth]{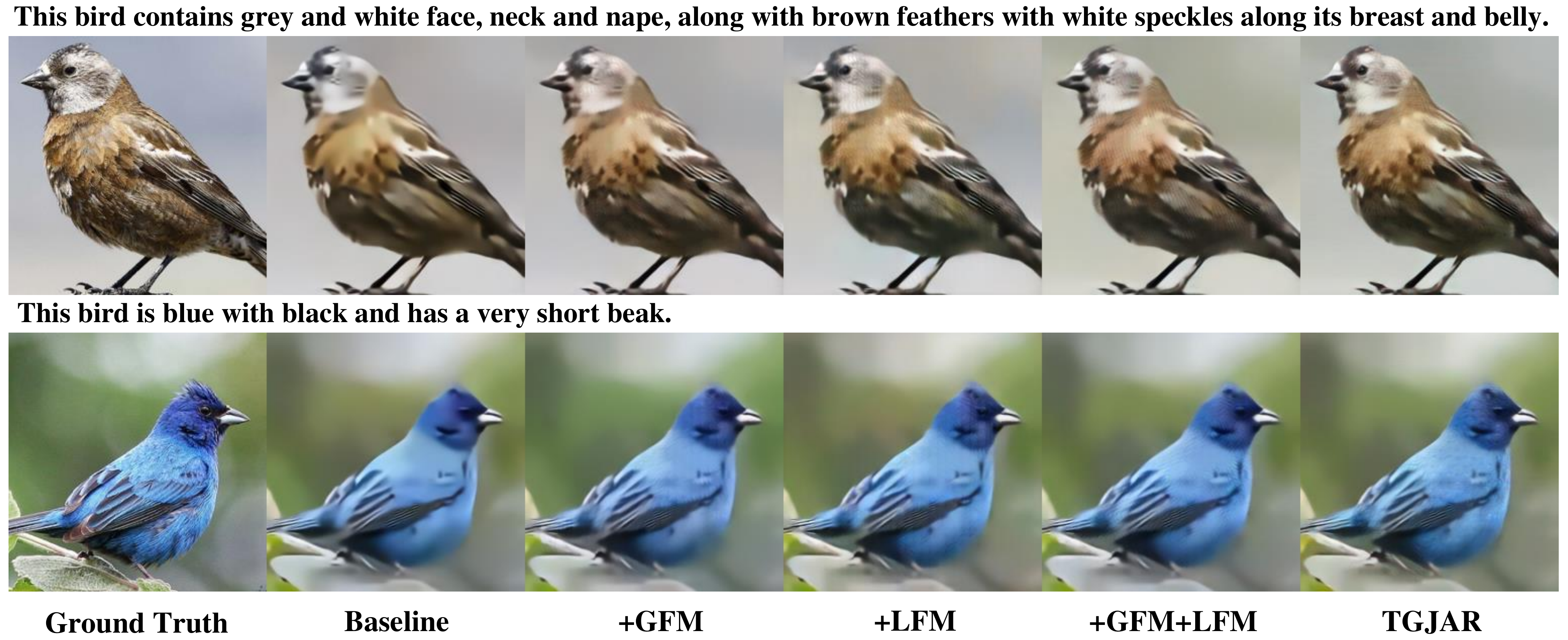}
	\end{center}
	\caption{Ablation study on CUB dataset. Better zoom in.}
	\label{fig:ablation}
\end{figure}

The qualitative comparisons of ablation study are shown in Fig.~\ref{fig:ablation}. By comparing the deblocking results, the final TGJAR produces the deblocking results with the highest subjective quality. In addition, we find that only applying the GAN design, the deblocking results are still a little blurry. By separately introducing LFM and GFM, the model can effectively remove the blocking artifacts, and greatly improve the subjective quality of the image. However, the deblocking results are still relatively blurry by only applying GFM, and some unnatural textures exist in the deblocking results only adopting LFM. Thus, using the two modules simultaneously can give full play to the advantages of both modules, and restrict the disadvantages of each module. Unfortunately, we find that there are still some unnatural textures exists in the deblocking results using both two modules. Recognizing this problem, we introduce contrast learning to obtain more real deblocking results with photo-realistic details.

\subsection{Text-guided Explorative Image Deblocking}
\begin{table}[t]
	\scriptsize
	\begin{center}
		\begin{tabular}{c|c|c}
			\hline
			\multirow{1}*{Dataset}&Method&LPIPS/FID/PI/NIQE\\
			\hline
			\multirow{2}*{COCO}&EDSR&0.458/165.3/7.63/9.19\\
			\cline{2-3}
			~&TGJAR&\textbf{0.269/~70.6~/5.91/7.68}\\
			\hline
			\multirow{2}*{Kodak }&EDSR&0.561/255.3/6.17/7.05\\
			\cline{2-3}
			~&TGJAR&\textbf{0.348/120.5/4.43/5.51}\\
			\hline
		\end{tabular}
	\end{center}
	\caption{Performance comparisons of our TGJAR and EDSR on COCO and Kodak datasets at QF = 1. Models are trained on training set of COCO, and tested on Kodak and 1000 images of testing set of COCO. The best results are boldfaced.}\label{tab:coco}
	
\end{table}

\begin{figure}[t]
	\begin{center}
		\includegraphics[width=1\linewidth]{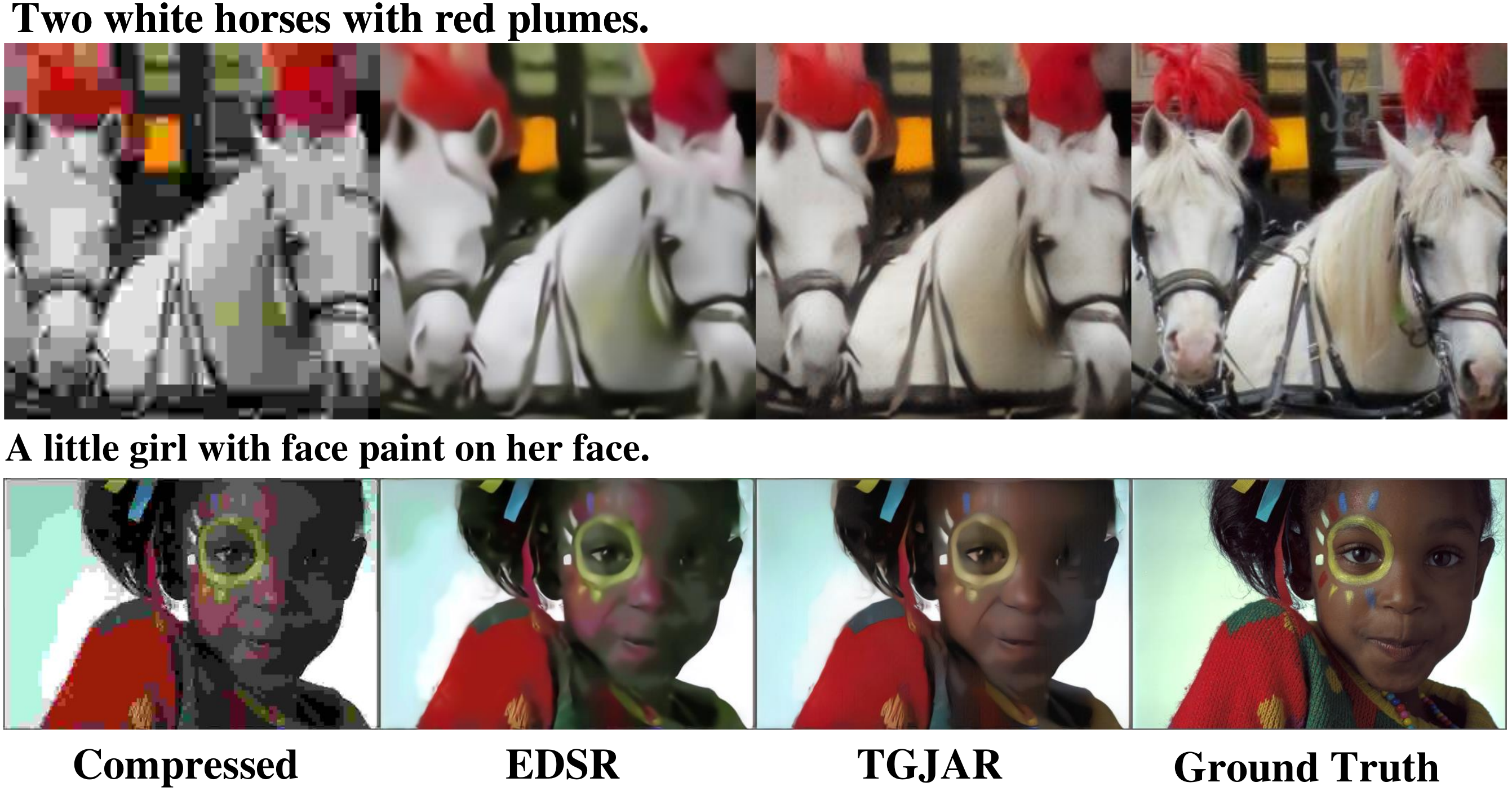}
	\end{center}
	\caption{Visual comparisons on COCO (first row) and Kodak (second row) at QF 1. Above each line of the images is the corresponding text description. Better zoom in. }
	\label{fig:coco}
\end{figure}


In order to explore more possibilities of our TGJAR, we also conduct experiments on COCO~\cite{lin2014microsoft} dataset with 80 types of objects at QF 1. To further verify the generalization performance of the algorithm, we also conduct the cross-dataset experiments on Kodak~\cite{franzen1999kodak}. Since Kodak does not have corresponding text descriptions, we consider using the image captions methods~\cite{wang2022ofa,li2020oscar,zhang2021vinvl} to generate texts for our experiments. Here, we use OFA~\cite{wang2022ofa} to produce the corresponding texts as an example. The generated texts are of high quality and highly semantically consistent with the images. We compare our TGJAR with EDSR, since both use the residual block as the main feature extraction module. The quantitative and qualitative experimental results are shown in Table \ref{tab:coco} and Fig.~\ref{fig:coco}. We can find that TGJAR can produce much better results compared with EDSR, which confirms the great potential of our TGJAR at extreme low-birate image compression artifacts reduction.

\subsection{Text-Guided Controllable Image Deblocking}
\begin{figure}[t]
	\begin{center}
		\includegraphics[width=1\linewidth]{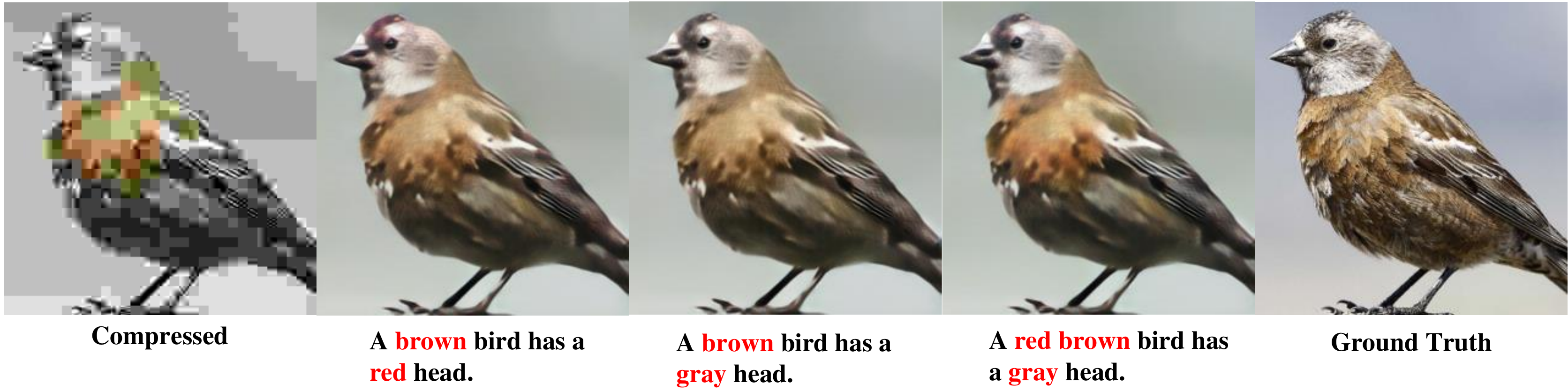}
	\end{center}
	\caption{Our results generated by using different text descriptions at QF 1. Better zoom in.}
	\label{fig:analysis}
\end{figure}
An experiment is further conducted to explore the effect of text guidance, that is, using different texts to assist in image deblocking. The visual results are shown in Fig.~\ref{fig:analysis}. Here, we modify the most obvious information, i.e., the color information. It can be found that the modification of the text descriptions does have an impact on the results, and the results are consistent with the semantics of the text descriptions. The previous single-modal methods can only enhance the image based on experience, but the multimodal algorithm can be personalized to enhance the image.

\subsection{Discussion}

Text-guided JPEG artifacts reduction needs to introduce a new modality, that is, the corresponding text description of the image. In this regard, we believe that there are three ways to obtain the texts. Firstly, many images on social media (e.g. Twitter, Facebook) are tagged with text descriptions consisted of rich attributes, which can be used for multimodal deblocking. Secondly, users can provide reasonable descriptions to enhance the image according to their imagination. As a result, the image will have a personalized enhancement effect. Thirdly, image caption algorithms can be used to obtain the text description corresponding to the image before the image is compressed, since original images are accessible during image compression. Then we can transmit texts as labels with the compressed images to the decoder side, and use these texts to enhance the compressed images. Note that the text occupies very few bits and can be transmitted to the decoder side at marginal bandwidth cost. Even for extreme low bitrate, i.e. JPEG quality factor set to 1, texts use less than one twenty-fifth of the bits used by images on datasets~\cite{CUB,flower}.

\section{Conclusion}
In this paper, we propose a text-guided generative adversarial network for JPEG artifacts reduction (TGJAR). To better fuse image and text information, we design two fusion modules, including the image-text global fusion module and the image-text local fusion module. These two modules fuse the global and local features of the image and text information from the global and local perspectives, respectively. Besides, to further improve the subjective quality of the deblocking results, a well-designed contrastive loss is built upon contrastive learning to constrain that the restored image is pulled to closer to the uncompressed image and pushed to far away from the compressed image in perceptual quality space. Experimental results demonstrate that TGJAR outperforms the test SOTA methods. Especially, even when QF=1, the TGJAR can produce visually pleasing results.


\bibliographystyle{named}
\bibliography{ijcai23}

\newpage
\appendix

\section{Appendix}

\subsection{Section A}
\begin{figure}[h]
	\begin{center}
		\includegraphics[width=1\linewidth]{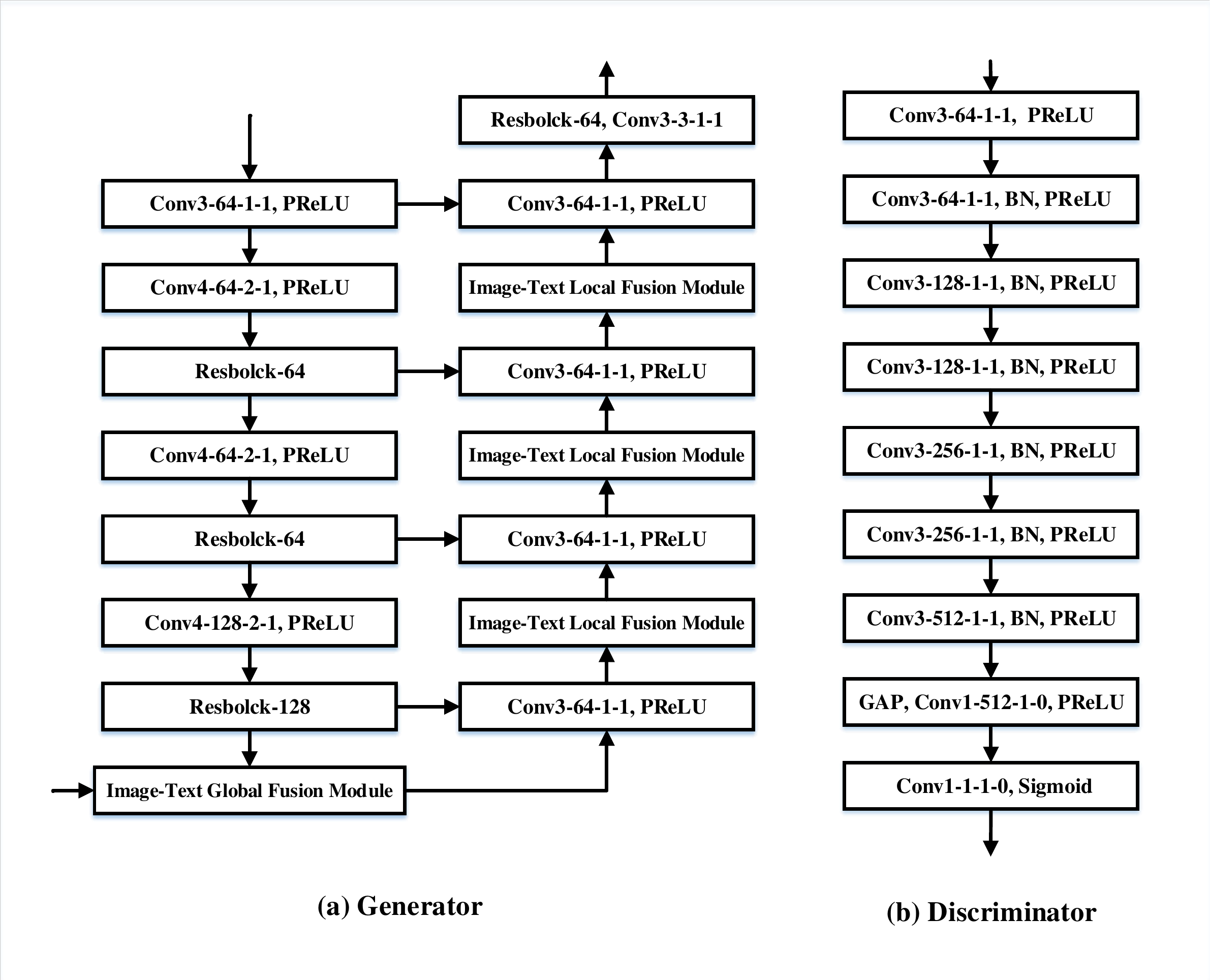}
	\end{center}
	\caption{Architectures of the generator and the discriminator in our model. (a) shows the detail parameters of the generator, and (b) represents the discriminator. The Conv3-64-1-1 means that the convolutional layer of size 3$\times$3, stride of 1, padding of 1 and kernel number of 64, and the resblock-64 means it is composed of two Conv3-64-1-1 and one PReLU. In addition, the BN represents the batch normalized layer, and the GAP represents the global average pooling layer.}
	\label{fig:generator-details}
\end{figure}

\textbf{Genrator} The details of the generator are shown in Fig. \ref{fig:generator-details} (a). The Conv3-64-1-1 means that the convolutional layer of size 3$\times$3, stride of 1, padding of 1 and kernel number of 64, and the resblock-64 means it is composed of two Conv3-64-1-1 and one PReLU. Note that in the generator, two kinds of image-text fusion modules including GFM and LFM, are designed to fuse image and text information from both global and local perspectives.

\textbf{Discriminator} Recognizing that GAN-based network has the ability to generate high subjective quality images in image generation tasks, we also adopt the GAN design in our TGJAR. The discriminator is used to determine whether the image is a restored image or an uncompressed image. The architecture of our discriminator is shown in Fig. \ref{fig:generator-details} (b). The BN represents the batch normalized layer, and the GAP represents the global average pooling layer.

\subsection{Section B}

\begin{figure*}[t]
	\begin{center}
		\includegraphics[width=0.95\linewidth]{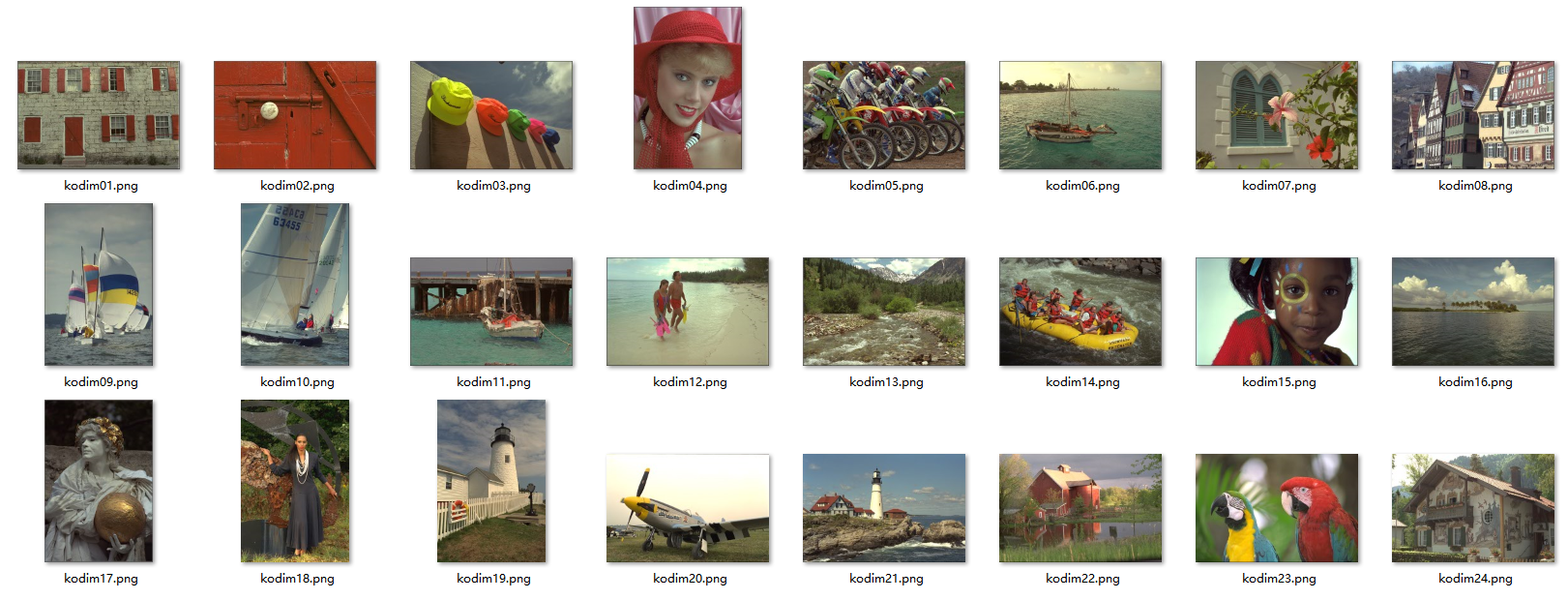}
	\end{center}
	\caption{Thumbnail image of all images on Kodak.}
	\label{fig:kodak}
\end{figure*}

\begin{table*}[t]
	\begin{center}
		\begin{tabular}{c|c|c|c|c|c}
			\hline
			\multirow{1}*{Method}&EDSR&RNAN&QGCN&FBCNN&TGJAR\\
			\hline
			Total Parameter (M)&19.4&8.9&12.3&71.9&6.9\\
			\hline
			Average Runtime (s)&0.178&0.505&0.605&0.262&0.087\\
			\hline
		\end{tabular}
	\end{center}
	\caption{\textbf{Comparison of Computational Cost.} The total parameter represents the parameter amount of EDSR, RNAN, QGCN and FBCNN. In TGJAR, the total parameter shows the sum of the parameter amount of the text encoder and generator. The average runtime represents the testing time of each image in CUB.  }\label{tab:cost}
\end{table*}

\textbf{The Generated Text Descriptions on Kodak}
We use the latest image caption model OFA~\cite{wang2022ofa} to generate the corresponding text descriptions for each images on Kodak~\cite{franzen1999kodak}. Here, we first provide a thumbnail image of all images on Kodak, which is shown in Fig.~\ref{fig:kodak}, and then report the generated texts as follows:
\begin{itemize}	
	\item kodak/kodim01.png: a stone building with red shuttered windows and a red door.
	\item kodak/kodim02.png: an old red door with a door knob on it.
	\item kodak/kodim03.png: a row of colorful hats hanging on a wooden wall.
	\item kodak/kodim04.png: a woman wearing a red hat and a red scarf and a pink shirt.
	\item kodak/kodim05.png: a group of dirt bike riders racing on a dirt track.
	\item kodak/kodim06.png: a small boat on the water in the ocean.
	\item kodak/kodim07.png: a hibiscus flower in front of a window.
	\item kodak/kodim08.png: a row of houses in a town with a mountain in the background.
	\item kodak/kodim09.png: a group of sailboats on the water with white sails.
	\item kodak/kodim10.png: a group of people on sailboats in the water.
	\item kodak/kodim11.png: a small boat in the water next to a dock.
	\item kodak/kodim12.png: a man and a woman walking in the water on the beach with flippers.
	\item kodak/kodim13.png: a river with mountains in the background.
	\item kodak/kodim14.png: a group of people on a raft in the river.
	\item kodak/kodim15.png: a little girl with face paint on her face.
	\item kodak/kodim16.png: a small island with palm trees in the middle of a body of water.
	\item kodak/kodim17.png: a statue of a woman holding a golden sphere.
	\item kodak/kodim18.png: a woman standing in front of a metal sculpture.
	\item kodak/kodim19.png: a white lighthouse with a white picket fence in front of it.
	\item kodak/kodim20.png: a small propeller plane parked on a field.
	\item kodak/kodim21.png: a lighthouse on a cliff with a body of water.
	\item kodak/kodim22.png: a red barn with a pond in front of it.
	\item kodak/kodim23.png: two colorful parrots standing next to each other.
	\item kodak/kodim24.png: a house with a mural on the side of it.	
\end{itemize}


We can find that the generated texts are of very high quality and are highly semantically consistent with the images. 

\subsection{Section C}

\textbf{Comparison of computational complexity.}
In practical applications, the parameters and runtime of the model are very important factors. Noted that the perceptual quality prediction module, the image encoder and the discriminator are only used in the training stage, and do not need to be used in the testing stage. Thus, we only calculate the sum of the parameters of the generator and the text encoder. As shown in the table~\ref{tab:cost}, it can be found that our TGJAR has the least parameters and takes the least runtime. Thanks to the assistance of text information, the proposed TGJAR not only runs fast, but also greatly improves the subjective quality of the deblocking results.

\begin{figure*}[t]
	\begin{center}
		\includegraphics[width=1\linewidth]{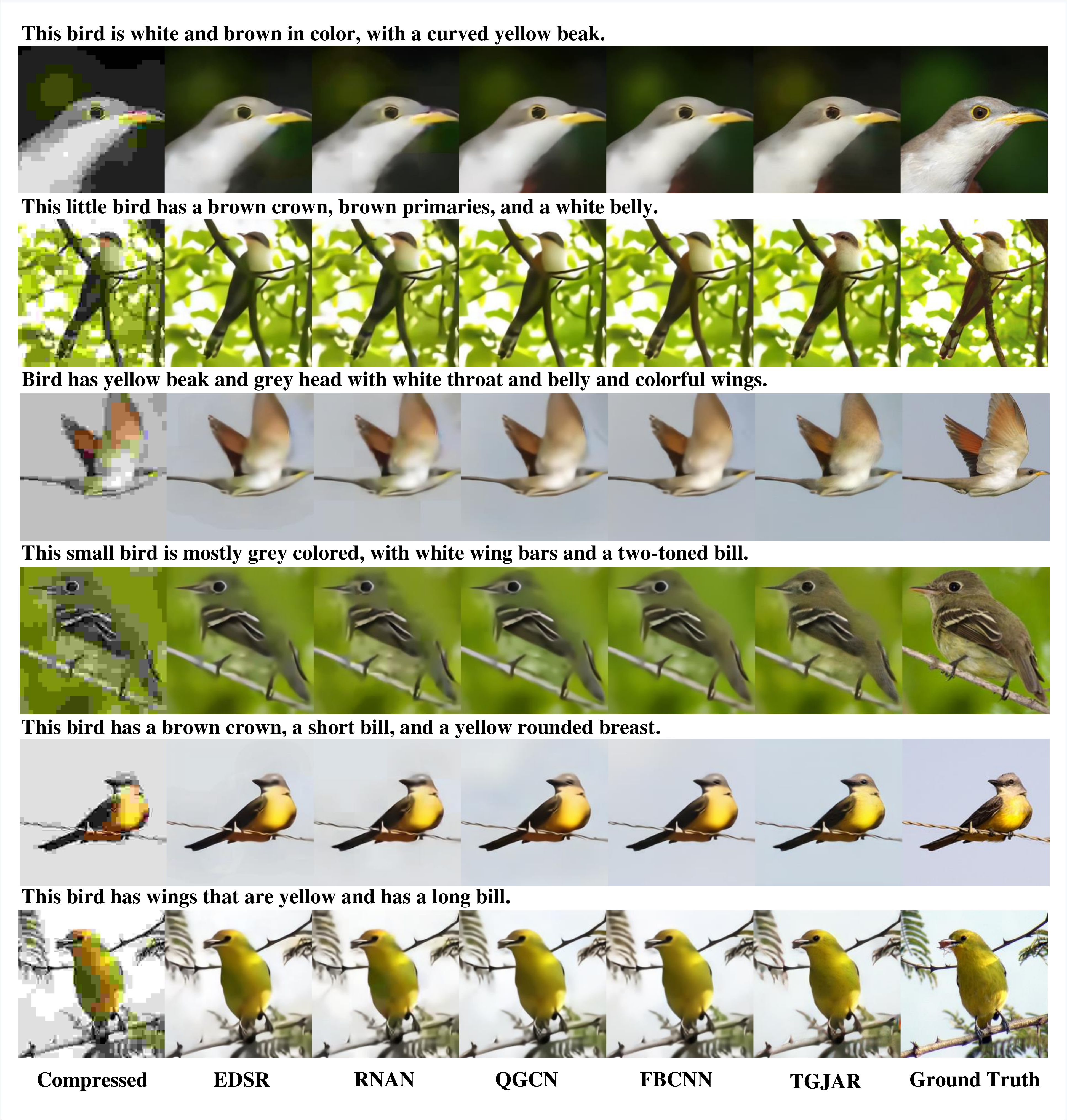}
	\end{center}
	\caption{Visual comparisons on \textbf{CUB} dataset of the proposed \textbf{TGJAR} and the state-of-the-art methods \textbf{EDSR}, \textbf{RNAN}, \textbf{QGCN} and \textbf{FBCNN}. The first column shows the compressed images, which are generated by compressing the ground truth (i.e. the seventh column) by using \textbf{QF 1}. The second, third, fourth, fifth and sixth columns show the deblocking results of these five algorithms, respectively. Better zoom in.}
	\label{fig:1}
\end{figure*}

\begin{figure*}[t]
	\begin{center}
		\includegraphics[width=1\linewidth]{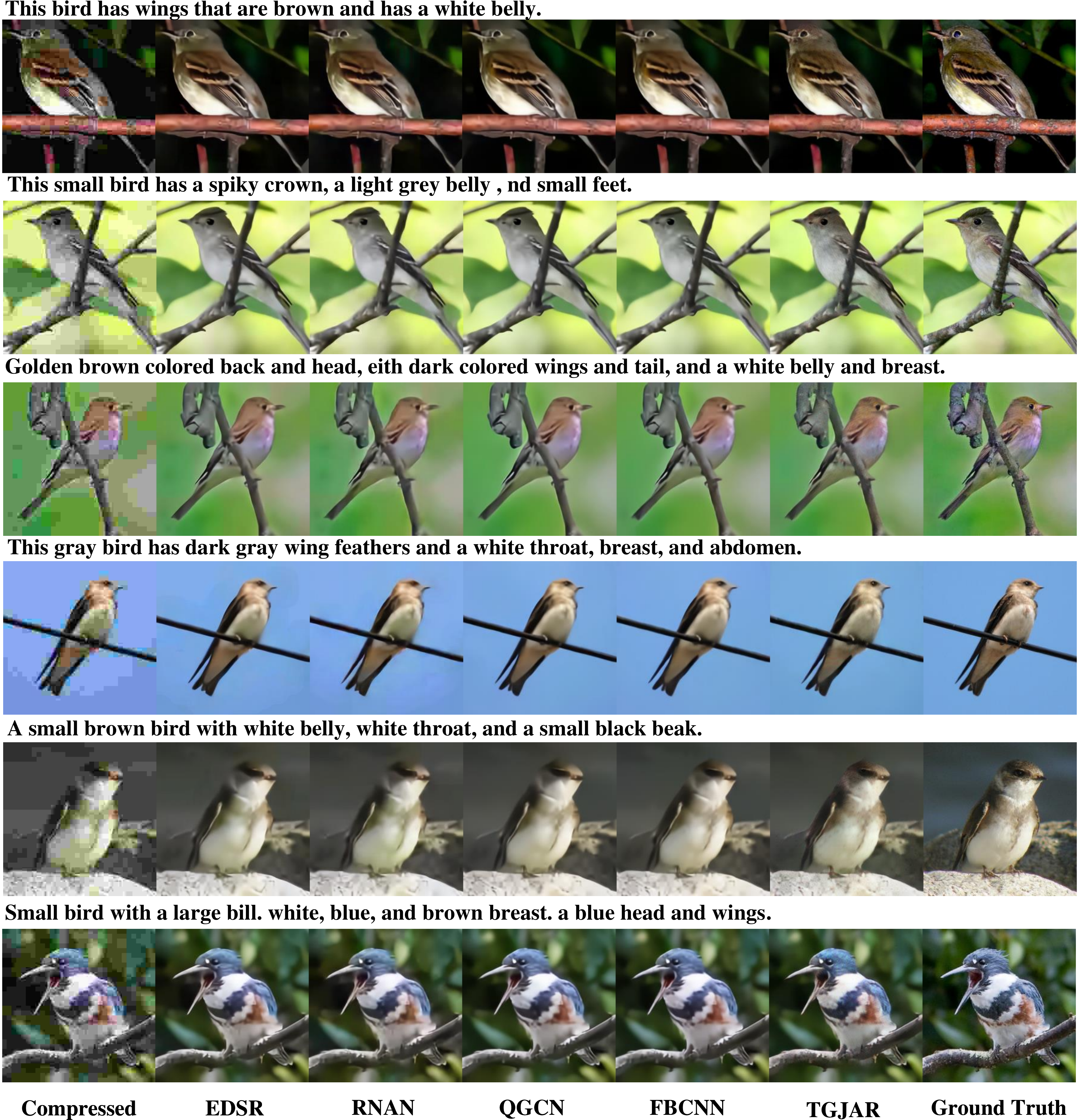}
	\end{center}
	\caption{Visual comparisons on \textbf{CUB} dataset of the proposed \textbf{TGJAR} and the state-of-the-art methods \textbf{EDSR}, \textbf{RNAN}, \textbf{QGCN} and \textbf{FBCNN}. The first column shows the compressed images, which are generated by compressing the ground truth (i.e. the seventh column) by using \textbf{QF 5}. The second, third, fourth, fifth and sixth columns show the deblocking results of these five algorithms, respectively. Better zoom in.}
	\label{fig:2}
\end{figure*}

\begin{figure*}[t]
	\begin{center}
		\includegraphics[width=1\linewidth]{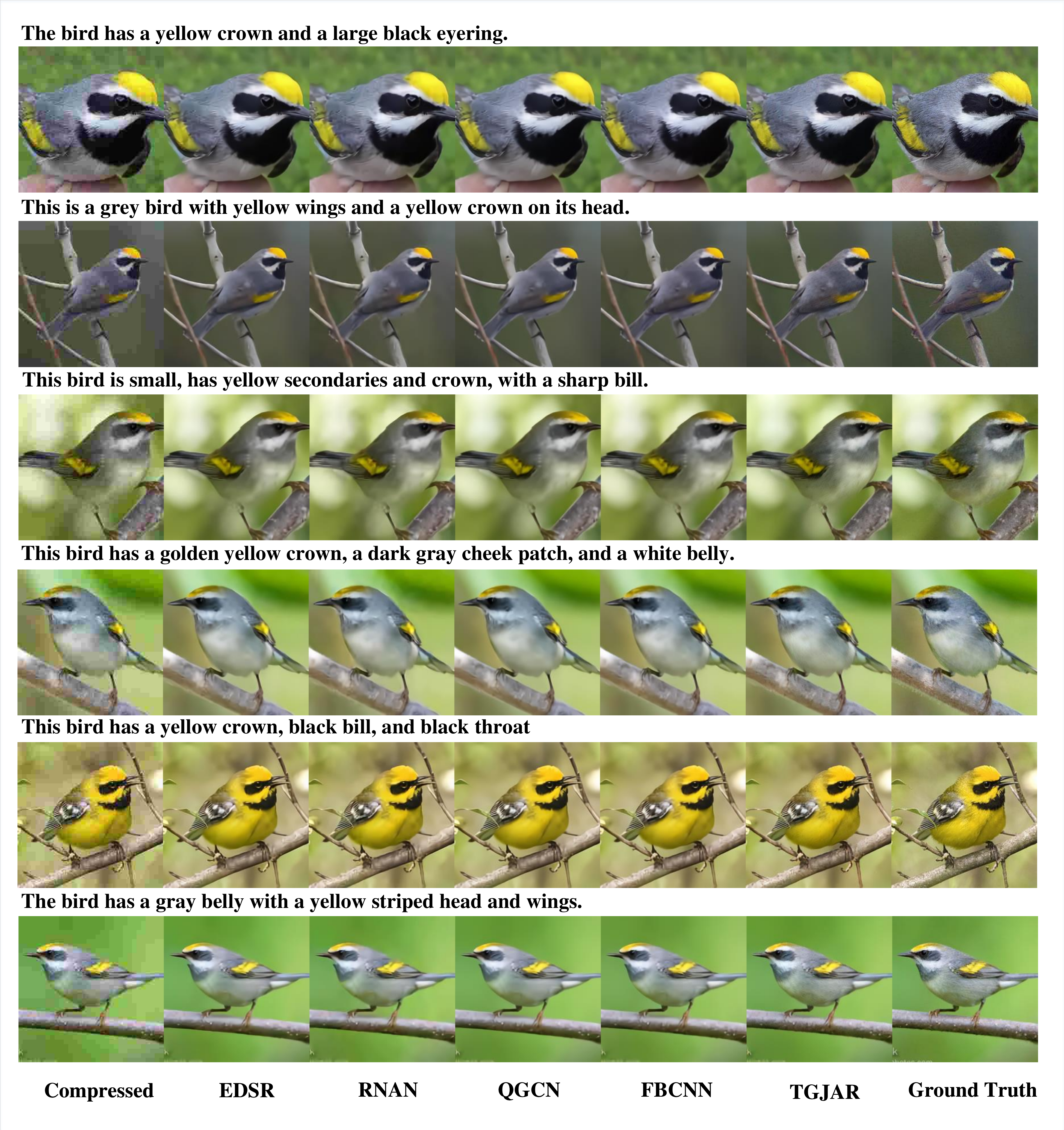}
	\end{center}
	\caption{Visual comparisons on \textbf{CUB} dataset of the proposed \textbf{TGJAR} and the state-of-the-art methods \textbf{EDSR}, \textbf{RNAN} and \textbf{QGCN}. The first column shows the compressed images, which are generated by compressing the ground truth (i.e. the sixth column) by using \textbf{QF 10}. The second, third, fourth and fifth columns show the deblocking results of these four algorithms, respectively. Above each line of the images is the corresponding text description. Better zoom in.}
	\label{fig:3}
\end{figure*}

\begin{figure*}[t]
	\begin{center}
		\includegraphics[width=1\linewidth]{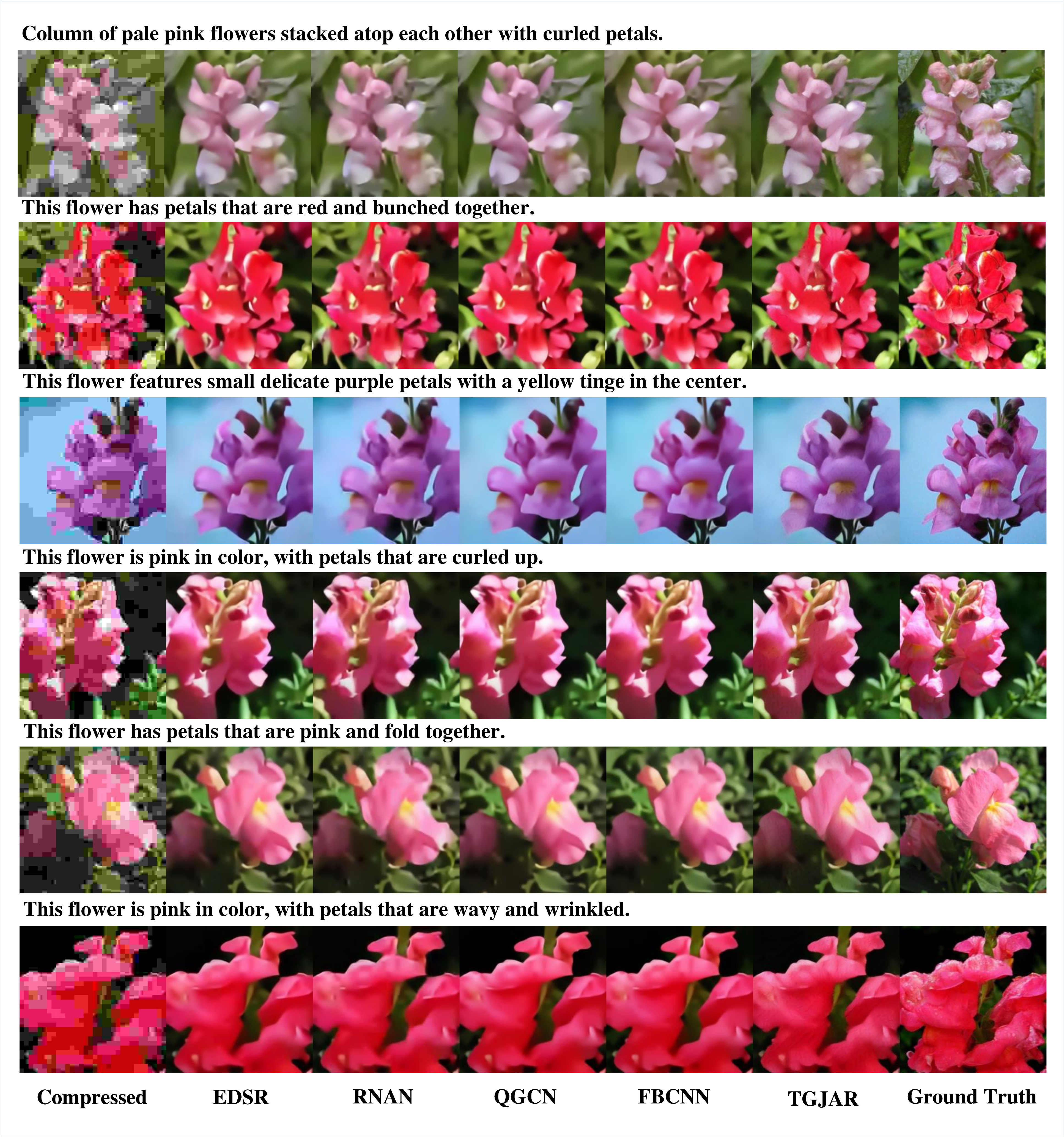}
	\end{center}
	\caption{Visual comparisons on \textbf{Oxford-102} dataset of the proposed \textbf{TGJAR} and the state-of-the-art methods \textbf{EDSR}, \textbf{RNAN}, \textbf{QGCN} and \textbf{FBCNN}. The first column shows the compressed images, which are generated by compressing the ground truth (i.e. the seventh column) by using \textbf{QF 1}. The second, third, fourth, fifth and sixth columns show the deblocking results of these five algorithms, respectively. Better zoom in.}
	\label{fig:4}
\end{figure*}

\begin{figure*}[t]
	\begin{center}
		\includegraphics[width=1\linewidth]{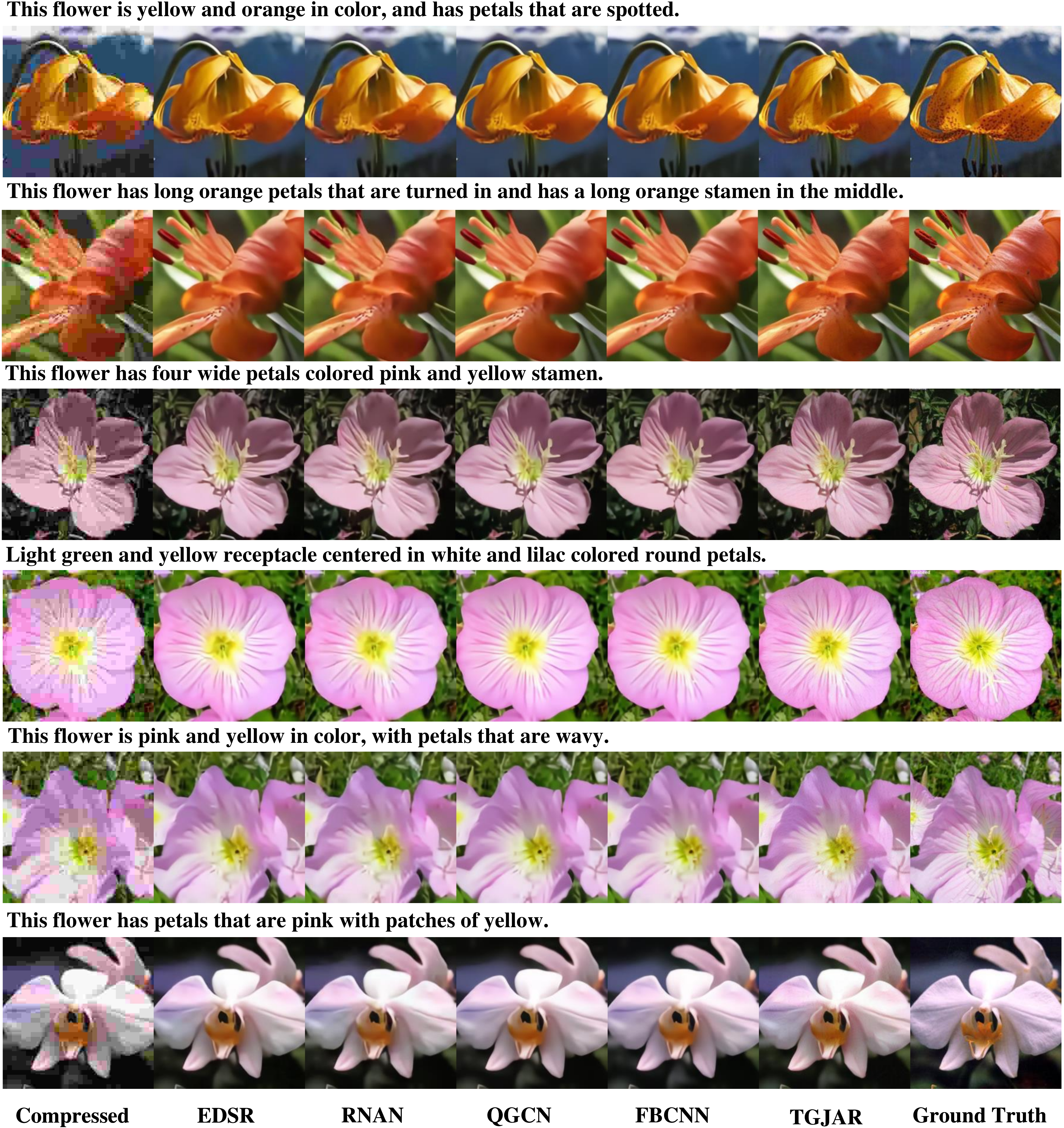}
	\end{center}
	\caption{Visual comparisons on \textbf{Oxford-102} dataset of the proposed \textbf{TGJAR} and the state-of-the-art methods \textbf{EDSR}, \textbf{RNAN}, \textbf{QGCN} and \textbf{FBCNN}. The first column shows the compressed images, which are generated by compressing the ground truth (i.e. the seventh column) by using \textbf{QF 5}. The second, third, fourth, fifth and sixth columns show the deblocking results of these five algorithms, respectively. Better zoom in.}
	\label{fig:5}
\end{figure*}

\begin{figure*}[t]
	\begin{center}
		\includegraphics[width=1\linewidth]{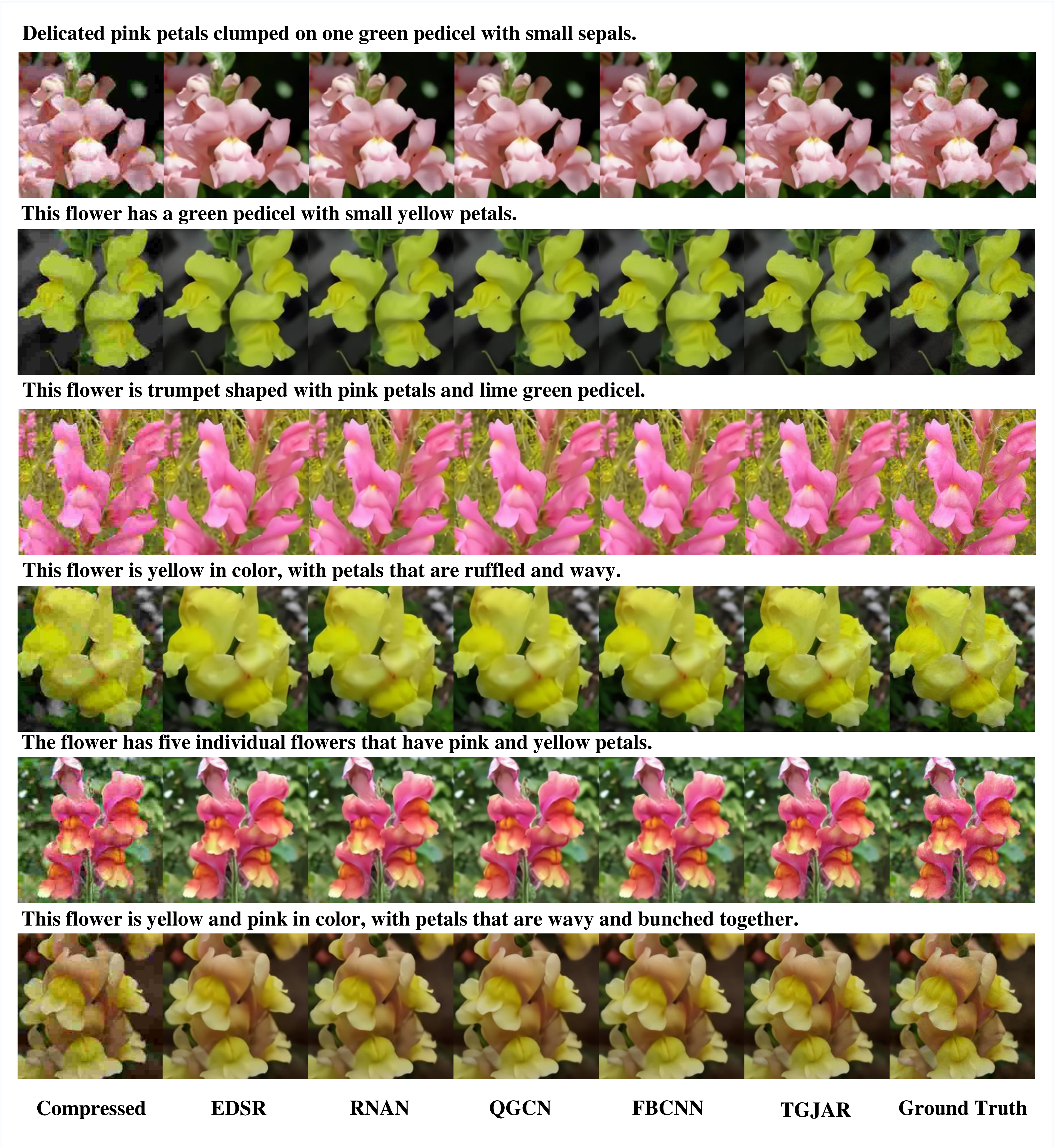}
	\end{center}
	\caption{Visual comparisons on \textbf{Oxford-102} dataset of the proposed \textbf{TGJAR} and the state-of-the-art methods \textbf{EDSR}, \textbf{RNAN}, \textbf{QGCN} and \textbf{FBCNN}. The first column shows the compressed images, which are generated by compressing the ground truth (i.e. the seventh column) by using \textbf{QF 10}. The second, third, fourth, fifth and sixth columns show the deblocking results of these five algorithms, respectively. Better zoom in.}
	\label{fig:6}
\end{figure*}

\textbf{More visualization results on CUB and Oxford-102}
We add a large number of visualization results. The image deblocking results of five algorithms, including the proposed TGJAR, EDSR~\cite{EDSR}, RNAN~\cite{zhang2019residual}, QGCN~\cite{li2020learning} and FBCNN~\cite{jiang2021towards}, are shown in Figs. \ref{fig:1}, \ref{fig:2}, \ref{fig:3}, \ref{fig:4}, \ref{fig:5} and \ref{fig:6}. Specifically, these results are obtained on CUB~\cite{CUB} and Oxford-102~\cite{flower} at each QF. As shown in Figs \ref{fig:1}-\ref{fig:6}, we can find that the subjective qualities of the results generated by TGJAR are higher than that of other models, and the proposed TGJAR faithfully reconstruct the deblocking results to the uncompressed images.

\begin{figure*}[t]
	\begin{center}
		\includegraphics[width=1\linewidth]{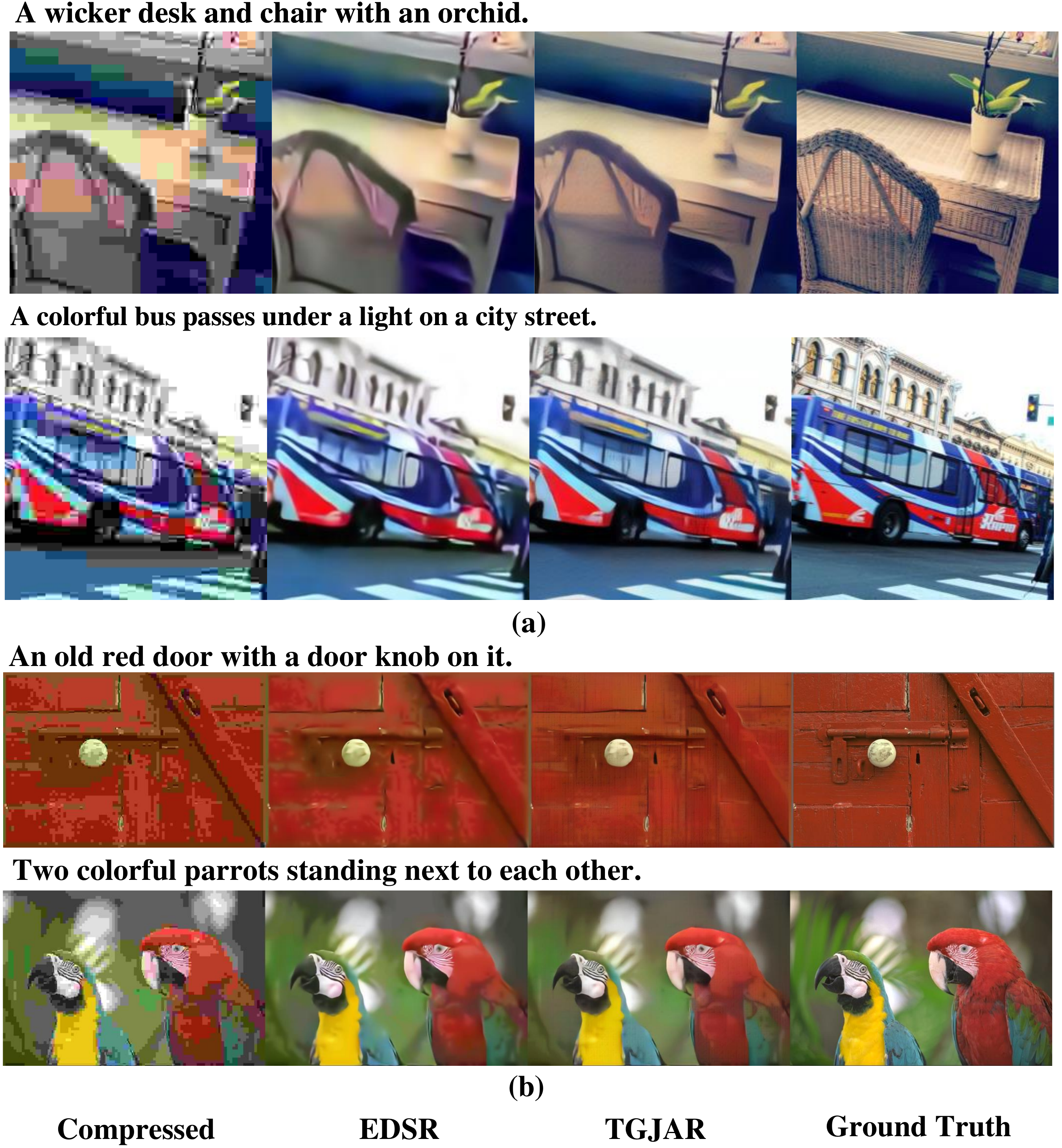}
	\end{center}
	\caption{Visual comparisons on \textbf{COCO (a)} and \textbf{Kodak (b)} of the proposed \textbf{TGJAR} and the state-of-the-art methods \textbf{EDSR}. The first column shows the compressed images, which are generated by compressing the ground truth (i.e. the fourth column) by using \textbf{QF 1}. The second and third columns show the deblocking results of these two algorithms, respectively. Above each line of the images is the corresponding text description. Better zoom in.}
	\label{fig:7}
\end{figure*}

\textbf{More visualization results on COCO and Kodak}
To explore more possibilities of our TGJAR, we also conduct experiments on COCO~\cite{lin2014microsoft} dataset at QF 1, which consists of 80 types of objects. We also conduct the cross-dataset experiments on Kodak~\cite{franzen1999kodak}, where the text descriptions of the images are generated by OFA~\cite{wang2022ofa}. We add a large number of visualization results. The reconstructed results of the proposed TGJAR and EDSR~\cite{EDSR} can be seen in Fig.~\ref{fig:7}. we can find that the subjective qualities of the results generated by our TGJAR are higher than that of EDSR.

\textbf{The impact of hyperparameters $\lambda$.}
We adopt the same GAN loss as in SRGAN (CVPR 2017), so we adopt the same hyper-parameters to set $\lambda_2$ and $\lambda_3$ to 1 and 0.001. $\lambda_1$ and $\lambda_4$ are determined empirically based on objective metrics after lots of experiments. Here, we provide the results with different $\lambda_1$ and $\lambda_4$ on CUB dataset at QF 1, as shown in Table~\ref{param}. Note that the last row of Table~\ref{param} shows the performance in the manuscript. Our hyperparameter setting achieves a trade-off between the four metrics, and obtains the best or the second best performance.

\begin{table}[t]
	
	\begin{center}
		\resizebox{0.5\textwidth}{!}
		{ 
			\begin{tabular}{cc||cc}
				\hline
				$\lambda_1$&LPIPS/FID/PI/NIQE&$\lambda_4$&LPIPS/FID/PI/NIQE  \\
				\hline
				0.1&\textcolor{red}{0.231}/22.4/\textcolor{red}{6.83}/\textcolor{red}{8.34}&      0.001&0.257/\textcolor{red}{17.5}/7.02/8.83  \\
				0.001&0.280/\textcolor{red}{16.5}/7.23/9.66&    0.0001&\textcolor{red}{0.246}/18.8/\textcolor{blue}{6.99}/\textcolor{blue}{8.80}  \\
				\hline
				0.01&\textcolor{blue}{0.249}/\textcolor{blue}{17.9}/\textcolor{blue}{6.95}/\textcolor{blue}{8.62}&    0.0005&\textcolor{blue}{0.249}/\textcolor{blue}{17.9}/\textcolor{red}{6.95}/\textcolor{red}{8.62} \\
				\hline
			\end{tabular}
		}
	\end{center}
	\caption{Performance comparisons with different $\lambda_1$ and $\lambda_4$. Red is the best and blue is the second best.}\label{param}
	
\end{table}

\textbf{The performance of our method with ($L_R+L_G+L_C$).}  We have done this experiment and the performance of our variation ($L_R + L_G + L_C$) (0.288/56.7/7.01/8.82) is worse than that of our final model (0.249/17.9/6.95/8.62) on CUB at QF 1.

\end{document}